\pdfoutput=1
\documentclass[10pt,twocolumn,times]{article}
\usepackage[pagenumbers]{cvpr} 

\usepackage{multirow}
\usepackage{makecell}
\usepackage[dvipsnames]{xcolor}
\usepackage{comment}
\usepackage{pifont}
\usepackage{colortbl}
\makeatletter
\newcommand*{\@rowstyle}{}

\newcommand*{\rowstyle}[1]{
  \gdef\@rowstyle{#1}%
  \@rowstyle\ignorespaces%
}

\newcolumntype{=}{
  >{\gdef\@rowstyle{}}%
}

\newcolumntype{+}{
  >{\@rowstyle}%
}
 \makeatother
\newcommand{\dataset}{R3DS\xspace}
\newcommand{\datasetfull}{Reality-linked 3D Scenes\xspace}
\newcommand{\task}{PanoSun\xspace}
\newcommand{\taskfull}{Panoramic Scene Understanding\xspace}
\newcommand{\datareal}{\textit{\dataset-real}\xspace}
\newcommand{\datasyn}{\textit{\dataset-syn}\xspace}
\newcommand{\datamix}{\textit{\dataset-mix}\xspace}
\newcommand{\datafull}{\textit{\dataset-full}\xspace}
\newcommand{\IG}{\textit{IG}\xspace}
\newcommand{\IGplusours}{\textit{IG+\dataset}\xspace}
\newcommand{\STD}{\textit{S3D}\xspace}

\newcommand{\best}[1]{\textbf{#1}}

\newcommand{\denselist}{\itemsep 0pt\parsep=0pt\partopsep 0pt}
\newcommand{\mypara}[1]{\noindent\textbf{#1}}
\newcommand{\cmark}{\textcolor{tblgreen}{\ding{51}}}%
\newcommand{\xmark}{\textcolor{red}{\ding{55}}}%

\definecolor{tblblue}{RGB}{31, 119, 180}
\definecolor{tblorange}{RGB}{255, 127, 14}
\definecolor{tblgreen}{RGB}{44, 160, 44}
\definecolor{tblred}{RGB}{214, 39, 40}

\definecolor{babyblue}{RGB}{208, 254, 254}
\definecolor{offwhite}{RGB}{255, 255, 228}

\newcommand{\R}[1]{{%
    \textbf{%
        \ifstrequal{#1}{1}{\textcolor{tblred}{R#1}}{%
        \ifstrequal{#1}{2}{\textcolor{tblblue}{R#1}}{%
        \ifstrequal{#1}{3}{\textcolor{tblorange}{R#1}}{%
                           \textcolor{tblgreen}{R#1}
        }}}%
    }%
}}

\definecolor{cvprblue}{rgb}{0.21,0.49,0.74}
\usepackage[pagebackref,breaklinks,colorlinks,citecolor=cvprblue]{hyperref}

\def\paperID{****} 
\def\confName{CVPR}
\def\confYear{2024}

\title{R3DS: Reality-linked 3D Scenes for Panoramic Scene Understanding}
\author{
Qirui Wu$^{1}$ \quad 
Sonia Raychaudhuri$^{1}$ \quad 
Daniel Ritchie$^{2}$ \quad
Manolis Savva$^{1}$ \quad 
Angel X. Chang$^{1,3}$\\
$^{1}$Simon Fraser University \quad
$^{2}$Brown University \quad
$^{3}$Alberta Machine Intelligence Institute (Amii)\\
{\tt\small \url{https://3dlg-hcvc.github.io/r3ds/}}
}

\begin{document}

\newcommand{\teaserfigure}{
\vspace{-2.75em}
\begin{center}
\captionsetup{type=figure}
\includegraphics[width=\textwidth]{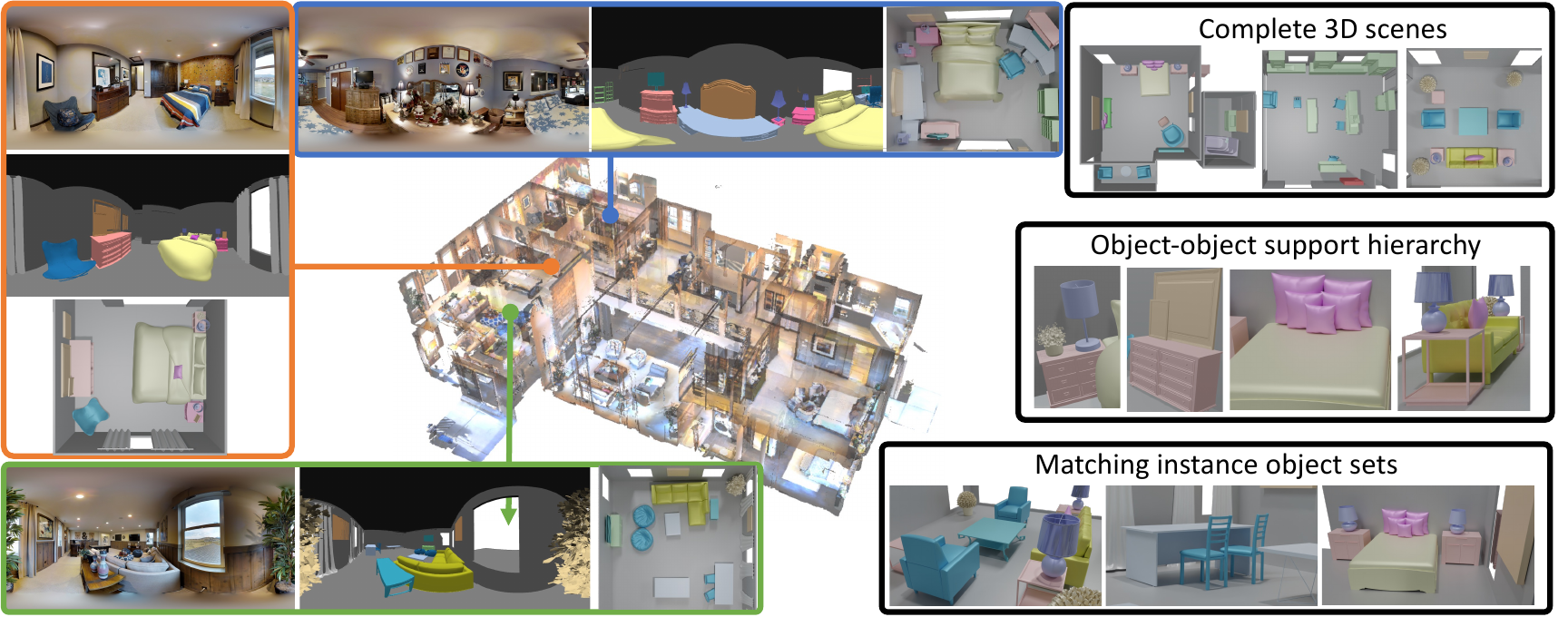}
\captionof{figure}{
\textbf{Left}: the \datasetfull dataset (\dataset) fills a gap between synthetic 3D scenes and reconstructions of real-world environments by providing 3D scene proxies linked to real-world panoramas from Matterport3D (three example panoramas and 3D scenes shown).
\textbf{Right}: our dataset contains scenes with higher density and completeness compared to prior datasets, and provides additional annotations such as object support (what objects or architectural elements support other objects), and matching object sets (e.g., pairs of the same nightstand). We use our dataset for the panoramic scene understanding task and demonstrate its value for research on room layout estimation, as well as 2D and 3D object detection.}
\label{fig:teaser}
\end{center}

\vspace{0.5em}
}

\twocolumn[{
\vspace*{-0.5cm}
\maketitle

\vspace{-2.75em}
\begin{center}
\captionsetup{type=figure}
\includegraphics[width=\textwidth]{fig/images/teaser.pdf}
\captionof{figure}{
\textbf{Left}: the \datasetfull dataset (\dataset) fills a gap between synthetic 3D scenes and reconstructions of real-world environments by providing 3D scene proxies linked to real-world panoramas from Matterport3D (three example panoramas and 3D scenes shown).
\textbf{Right}: our dataset contains scenes with higher density and completeness compared to prior datasets, and provides additional annotations such as object support (what objects or architectural elements support other objects), and matching object sets (e.g., pairs of the same nightstand). We use our dataset for the panoramic scene understanding task and demonstrate its value for research on room layout estimation, as well as 2D and 3D object detection.}
\label{fig:teaser}
\end{center}

\vspace{0.5em}

}]



\begin{abstract}
We introduce the \datasetfull (\dataset) dataset of synthetic 3D scenes mirroring the real-world scene arrangements from Matterport3D panoramas.
Compared to prior work, \dataset has more complete and densely populated scenes with objects linked to real-world observations in panoramas.
\dataset also provides an object support hierarchy, and matching object sets (e.g., same chairs around a dining table) for each scene.
Overall, \dataset contains 19K objects represented by 3,784 distinct CAD models from over 100 object categories.
We demonstrate the effectiveness of \dataset on the \taskfull task.
We find that:
1) training on \dataset enables better generalization;
2) support relation prediction trained with \dataset improves performance compared to heuristically calculated support; and
3) \dataset offers a challenging benchmark for future work on panoramic scene understanding.
\end{abstract}    
\section{Introduction}

\begin{figure*}[t]
\centering
\includegraphics[width=\linewidth]{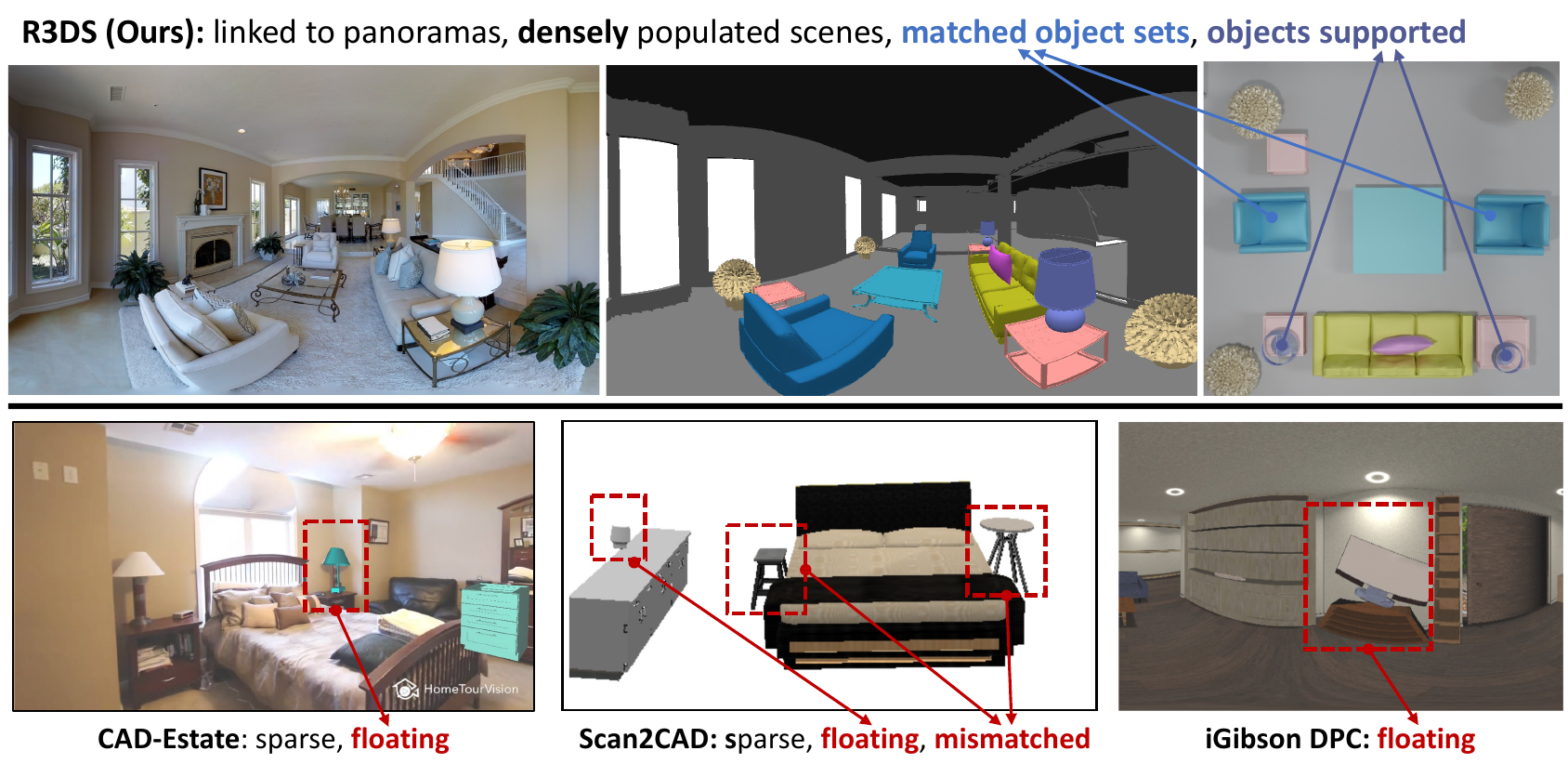}
\caption{
\textbf{Dataset comparison.} (Top) shows different views of a scene annotated in \dataset. Comparison with previous datasets (bottom) shows (1) \dataset has more complete scenes than the previous datasets; 
(2) Objects in \dataset are properly supported by either architecture or other objects unlike the others (e.g. floating objects with no proper support);
(3)~\dataset is annotated using the same 3D model for objects arranged together (chairs by the dining table, couches arranged together).
} 
\label{fig:dataset_comparison}
\end{figure*}

Datasets of 3D indoor environments are increasingly used for research on scene understanding~\cite{chang2017matterport3d,avetisyan2019scan2cad,zhang2021deeppanocontext}, embodied AI~\cite{batra2020rearrangement,szot2021habitat,shen2021igibson}, and scene generation~\cite{wang2019planit,paschalidou2021atiss}.
There are two strategies for constructing 3D scene datasets: reconstruction of real-world spaces~\cite{chang2017matterport3d}, or authoring scenes using synthetic 3D objects~\cite{fu20203dfront} (CAD
models).
Reconstruction captures real spaces but is hard to scale, and the resulting scenes exhibit imperfections and artifacts.
On the other hand, synthetic 3D scenes are complete and easy to manipulate but often do not match the statistics of real-world spaces and are artificially ``clean''.
Moreover, both strategies are time-consuming and require expertise.  
There have been some attempts to create ``synthetic'' replicas of real environments by matching CAD models to objects in scans~\cite{szot2021habitat,shen2021igibson}.
These efforts have been limited in scale and often result in partial and sparsely populated synthetic counterparts of the real environments.


We design a framework that allows users to create 3D scenes from RGB panoramas and use it to create \textbf{\dataset}: a dataset of `\textbf{R}eality-linked' \textbf{3D} \textbf{S}cenes.
Each 3D scene in our dataset is a complete proxy of an environment from the Matterport3D~\cite{chang2017matterport3d} dataset, representing both the 3D architecture and the objects.
Thus, each scene is linked to a real space, with correspondences established between panorama observations of each object and the synthetic object.
These reality-linked scenes reflect denser real-world arrangements of objects.

The use of panoramas for reference is advantageous compared to either perspective images or 3D reconstructions.
Panoramas are not limited by the field of view unlike perspective images, enabling more complete 3D synthetic proxies.
Panoramas also better capture relatively small objects and objects with challenging materials or illumination conditions compared to reconstructions.
Additionally, there is a scarcity of synthetic 3D scenes coupled with real-world panoramas, with only one relatively small algorithmically constructed dataset provided by \citet{zhang2021deeppanocontext} being available to the community.

Compared to prior efforts such as Scan2CAD~\cite{avetisyan2019scan2cad} and CAD-Estate~\cite{maninis2023cad}, our dataset provides more complete scenes, with salient observed objects being captured in the layout.
Moreover, we provide a support hierarchy defining what objects are placed on other objects and specify sets of identical objects such as dining chairs around a table, allowing for creating realistic variations of the scene by swapping the entire set to a different chair design.

We demonstrate the value of our dataset by using it for the \taskfull task.
We show that leveraging the denser layouts and support hierarchy information in our scenes leads to improved object detection performance and better generalization compared to training using other datasets previously used for this task.
In summary, we make the following contributions:
\begin{itemize}\denselist
\item We design a framework for efficient construction of synthetic scenes from real panoramas and use it to create \dataset: a dataset of reality-linked 3D scenes.
\item \dataset provides more complete and realistic scenes with correspondences between real and synthetic objects, and object-object support relations.
\item We show that the more complete layouts and support relations in our dataset enable better performance and generalization in the \taskfull task, and that our dataset offers a challenging benchmark for future work in scene understanding.
\end{itemize}

\section{Related Work}

\mypara{3D scene datasets.} 
A spectrum of scene datasets have been used for scene understanding tasks.
One type provides annotated 3D reconstructions of real scenes based on RGB-D videos~\cite{hua2016scenenn,dai2017scannet,chang2017matterport3d,straub2019replica,ramakrishnan2021hm3d,yadav2022hm3dsem}.
These datasets are usually subject to the limitations of RGB-D reconstruction, typically containing noise, artifacts such as holes, and poor reconstructions of thin structures, shiny objects, or light sources.
Another type of 3D datasets is authored by manually designing 3D object assets~\cite{fu20203dfuture,collins2022abo} and inserting them into synthetic 3D scenes~\cite{fu20203dfront}.
However, such datasets lack the realism of real-world reconstructions and demand expert knowledge, making them expensive to create.
A third, hybrid approach which is closest to our work creates 3D scene datasets by aligning existing object CAD models to real world data.


\mypara{Datasets that align CAD models to real world.}
There have been a number of recent efforts in aligning CAD models with real-world data.  
Prior work~\cite{sun2018pix3d, lim2013parsing,xiang2016objectnet3d} has annotated object images with 3D models, typically using keypoint correspondences to perspective images.
These perspective images usually do not depict a complete scene; they typically focus on one or two objects and are limited in field of view, resulting in a sparse proxy of the real scene.

Another line of work aligns 3D CAD models to RGB-D scans either through annotation as in Scan2CAD~\cite{avetisyan2019scan2cad}, or automated heuristics as in iGibson~\cite{shen2021igibson}.
OpenRooms~\cite{li2021openrooms} extends Scan2CAD~\cite{avetisyan2019scan2cad} with photorealistic material annotations and focuses on inverse rendering tasks. 
Conceptually, these allow for more complete synthetic scene proxies.
However, statistics from these datasets show that they are still relatively sparse (see \cref{tab:data_compare_stats}).
In addition, the poor quality of reconstruction makes aligning CAD models challenging without referring to the original RGB images.  
A prominent exception is Replica~\cite{straub2019replica} which has fairly high-quality reconstructions and the artist-created Replica-CAD~\cite{szot2021habitat}.
However, creating such high quality ``replicas'' is labor intensive and costly.
\citet{szot2021habitat} report 900+ work hours required to model approximately 90 objects, resulting in a dataset of limited scale with 105 different layouts of what is effectively a single room.

More recently, \citet{maninis2023cad} introduced CAD-Estate, which aligns CAD models to RGB videos for over 19K spaces.
Because the data is based on monocular video, the coverage of the spaces is incomplete.
In addition, the annotation is relatively sparse, with an average of only 6 objects per scene.


\mypara{Datasets for panoramic scene understanding.}
There have been relatively few datasets introduced for \taskfull~\cite{zhang2014panocontext,zhang2021deeppanocontext,dong2023panocontext}.
In the initial PanoContext dataset~\cite{zhang2014panocontext}, the data did not have aligned CAD models and only included object cuboids.
The ground truth data was collected on 2D panorama images by annotating visible cuboid vertices; 
3D cuboids were obtained by minimizing the re-projection error from the annotated 2D vertices.
Moreover, these 3D cuboids and the room layout are obtained with the assumption that the room layout is a cuboid and that the objects are vertically aligned. 
Thus, the resulting object layout may deviate from the real arrangement of objects.
More recently, datasets for \taskfull have been built by taking 3D scans, aligning CAD objects to them, and then generating panoramas~\cite{zhang2021deeppanocontext,dong2023panocontext}.
Compared to these datasets, our \dataset is manually curated for a larger number of distinct regions and provides support hierarchy and matching object set annotations.

\begin{table*}[t]
    \centering
    \resizebox{\linewidth}{!}{
    \begin{tabular}{@{}=l+l+c+c|+r+r|+r+r+r+r+r|+c+c@{}}
        \toprule
        Dataset  & Source & CAD Alignment & Type & Houses/Rooms & Panos & \#CAD &\#Objects &\#Cat & Ave Obj & Ave Cat  & Sup & Match
         \\
        \midrule
        Scan2CAD~\cite{avetisyan2019scan2cad} & ScanNet~\cite{dai2017scannet} & Annotator & scan & - / 1506 & \xmark & 3,049 & 14,225 & 35 & 9.4 & 4.1 & \xmark & \xmark \\

        OpenRooms~\cite{li2021openrooms} & ScanNet~\cite{dai2017scannet} & Scan2CAD~\cite{avetisyan2019scan2cad} & scan & - / 1288 & \xmark & 2,651 & 16,014 & 38 & 12.4 & 6.3 & \xmark & \xmark \\

        ReplicaCAD~\cite{szot2021habitat} & Replica~\cite{straub2019replica} & Artist recreation & scan & - / 105$^*$ & \xmark &92 & 2,293 &44 &21.8 &14.4 & \xmark & \xmark \\
        
        CAD-Estate~\cite{maninis2023cad} & RealEstate10K~\cite{zhou2018stereo} & Annotator & video & 19,512 & \xmark &12,024 &100,882 &49 &6.3 &3.4 & \xmark & \xmark \\

         
        \midrule
        
        \rowstyle{\color{gray}}
        Replica-Pano~\cite{dong2023panocontext} & Replica~\cite{straub2019replica} & Heuristic & pano & - / 27 & 2700 & - & - & 25 & - & - & \xmark & \xmark \\
        iGibson-DPC~\cite{zhang2021deeppanocontext} &  iGibson~\cite{shen2021igibson} & Heuristic & pano & 15 / 100 &1500 &500 &26,998 &57 &17.9 &10.2 & \xmark & \xmark \\

        \rowcolor{offwhite}
        \textbf{\dataset (Ours)} & Matterport3D~\cite{chang2017matterport3d} & Annotator & pano & 20 / 370 & 842 &3,784 & 19,050 & 110 & 22.9 &10.4 & \cmark & \cmark \\

         \bottomrule
    \end{tabular}}
    \caption{\textbf{Comparison with 3D indoor scene datasets aligned with real-world images, videos, or scans.}
    Our \dataset dataset contains more densely populated annotations compared to other datasets, with objects from 110 different categories.  
    We report the unique models (\#CAD), object categories (\#Cat), object instances (\#Objects) as well as average number of objects and object categories per annotation.
    For Scan2Cad~\cite{avetisyan2019scan2cad} and ReplicaCAD~\cite{szot2021habitat} the average is 
    per scan.
    Note that ReplicaCAD consists of 105 different layouts (arrangements) of effectively one room.
    CAD-Estate has partial views into 19K spaces, many of which are 1-2 rooms.
    Of the datasets used for panoramic scene understanding, our \dataset dataset covers more rooms with 842 panoramas over 22 room types.
    Replica-Pano (in gray) was not released, so we report statistics from the paper.
    Our annotations per panorama are more complete and our dataset has both support relations (Sup) and matching object instance sets (Match).
    }
    \label{tab:data_compare_stats}
\end{table*}

\begin{figure}
    \centering
    \includegraphics[width=\linewidth]{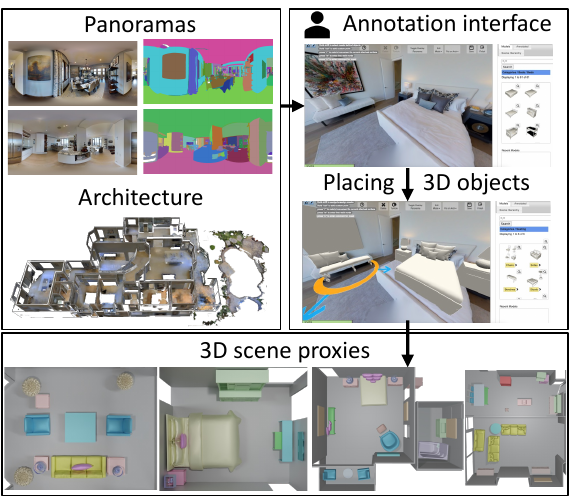}
    \caption{
    \textbf{\dataset annotation pipeline}.
    Annotators see an empty scene (architecture only).
    They then insert and manipulate 3D object models from a panorama viewpoint to create a populated 3D scene proxy corresponding to the panorama.}
    \label{fig:rlsd_system}
\end{figure}

\section{The \dataset Dataset}

We describe the construction of the \dataset dataset and present a statistical analysis of the scenes it contains.
Compared to previous datasets~\cite{maninis2023cad,avetisyan2019scan2cad, zhang2021deeppanocontext} (\cref{fig:dataset_comparison}), our scenes are more densely populated, and objects are annotated with a hierarchy of support relations. 
Moreover, our dataset specifies matching object instances in furniture arrangements.
\Cref{fig:teaser} shows example annotations from our dataset.

\subsection{Dataset construction}
\label{sec:annotation-interface}

We developed a 3D annotation interface  (\cref{fig:rlsd_system}) showing a panorama of a room from Matterport3D and allowing users to insert 3D CAD objects into a 3D scene which is \emph{visually overlaid} on the panorama.
The 3D scene is initially empty, consisting only of 3D architectural geometry which specifies the walls, floor, ceiling as well as the placement of openings (e.g. doors, windows, and other openings) on the walls.
We create this 3D architecture by taking 20 houses from Matterport3D, constructing an initial architecture based on the region and object annotations for the windows and doors, and manually refining the placement of walls and openings.
By combining panoramas and 3D architectures, users can see through openings and annotate objects located in other rooms.

We ask annotators to select and place 3D object models to best match the panoramic image.
We use CAD models from Wayfair~\cite{wayfair2016models} and ShapeNet~\cite{chang2015shapenet} models collected from 3D Trimble warehouse.
Wayfair provides a large collection of furniture CAD models that match real-world products and are sized based on real-world dimensions.
However, it does not include bathroom fittings, electronic equipment and kitchen appliances, for which we manually scale and align CAD models from ShapeNet.
Compared with ShapeNetCore, the CAD models we use are already sized to real-world sizes (instead of normalized to a unit cube).

To assist the annotators, we provide segmented masks of objects visible in the panorama.
Since Matterport3D has annotated 3D object masks on the scans we use those annotations, but it is also possible to run an instance segmentation on the panorama.
When the user clicks one of these masks, a search panel automatically opens and shows objects matching the clicked mask category label.  
For each mask, the annotator selects a matching object and positions and aligns it to match the mask.
Annotators are instructed to choose objects which match the \emph{shape} of the corresponding object in the panorama (rather than its color or texture).
To help annotators focus on shape, we render all 3D objects in a neutral gray color.
Annotators are also explicitly asked to select the same 3D asset for objects that should be the same; our interface provides a list of recently selected assets to make this process easier.
In addition, annotators are instructed to add annotations for any objects that are not segmented (due to errors in Matterport3D) through simple clicks.
These additional objects provide a more complete annotation that covers poorly reconstructed objects such as glass tables, lamps, and other small objects.
The interface enforces that each object is placed on a support surface (either an architecture element or another object).
The annotator can review their work by toggling off the panorama overlay or by switching to a perspective view of the 3D scene.
For more annotated scene examples and details on the annotation process please refer to the supplement. 

\begin{figure}[t]
\centering
\includegraphics[width=\linewidth]{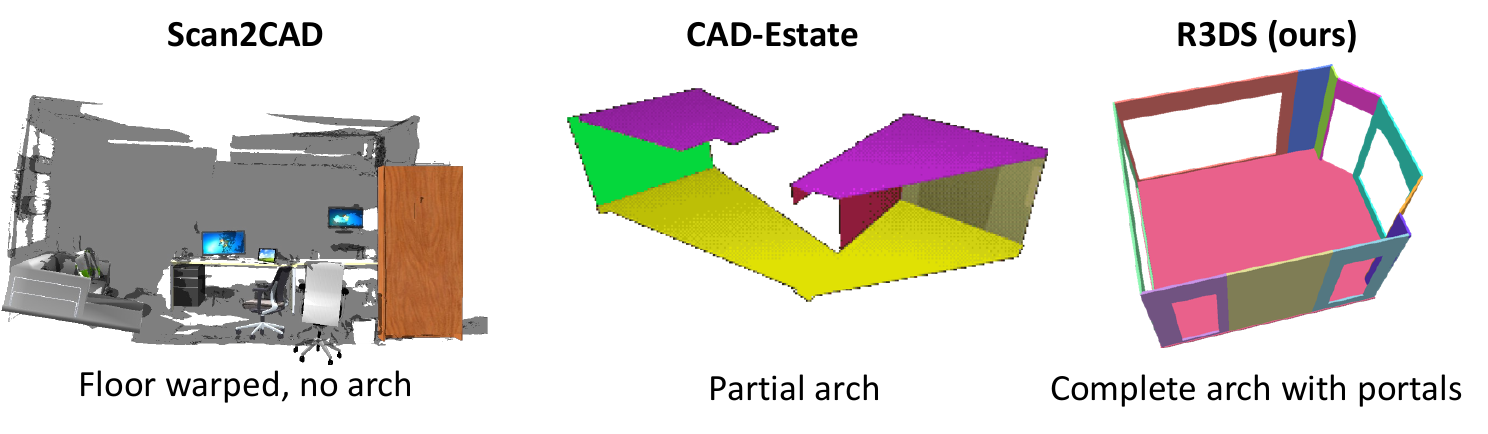}
\vspace{-10pt}
\caption{
\textbf{Architecture comparison.}
Compared to Scan2CAD (no architecture) and CAD-Estate (partial architecture), \dataset provides complete architecture with door/window portals. 
}
\label{fig:arch_comparison}
\end{figure}

\subsection{Dataset analysis and statistics}

We collect annotations for 20 Matterport3D houses with 808 panoramas in total.
We discard panoramas taken on stairs or outside a house, since they have a limited number of objects that can be placed.
After filtering we have 769 panoramas for our analysis and experiments.
For 73 panoramas, we collect two sets of annotations for each, to obtain a total of 842 annotated object arrangements across 22 different Matterport3D region types.
The panoramas with two annotations serve as a test of annotator consistency and add diversity.
In total, \dataset contains 19,050 object instances from 3,784 unique 3D CAD models spanning over 110 fine-grained object categories.
\Cref{tab:data_compare_stats} shows a comparison of overall statistics with previous 3D indoor scene datasets.
See the supplement for more statistics about annotated objects.

Compared to prior datasets that align CAD models to real-world scenes, \dataset is more complete, providing annotated object support hierarchies and matching object instances.
CAD-Estate~\cite{maninis2023cad} annotates RGB videos with 3D objects and architectural room layouts.
However, the architecture is partial as the videos have limited view (\cref{fig:arch_comparison}), and not all objects in the scenes are annotated (\cref{fig:dataset_comparison}).
This results in annotations of objects floating mid-air and not properly supported (e.g. lamp \cref{fig:dataset_comparison}).
Scan2CAD~\cite{avetisyan2019scan2cad} also lacks support structure (e.g., lamp not supported by cabinet in \cref{fig:dataset_comparison}).
In addition, because Scan2CAD does not provide clean 3D architecture on which objects are placed (\cref{fig:arch_comparison}), objects on the floor are not always placed such that their bottom face is parallel with a horizontal plane.
In contrast, our \dataset scenes have an accurate support hierarchy by construction.
OpenRooms~\cite{li2021openrooms} augments Scan2CAD with room layouts representing the architecture.
However, the architecture in \dataset is more complex and realistic, especially due to inclusion of more doors (1.92 doors per room in \dataset vs 0.67 in OpenRooms).

Evaluation of CAD object annotation quality is non-trivial as the `ground truth' from the semantically annotated 3D reconstructions is itself imperfect.
We measured how closely our annotated CAD objects conform to the real objects using the average 2D IoU between CAD object mask and ground truth 2D mask.
\dataset is at 42.6\% vs 38.5\% for Scan2CAD, across 8 common object categories (bed, sofa, chair, cabinet, tv/monitor, table, shelving, bathtub).

Of the datasets previously used for \taskfull, Replica-Pano~\cite{dong2023panocontext} has not been released, and  iGibson-DPC~\cite{zhang2021deeppanocontext} is the only dataset with synthetic panoramic images annotated with 3D objects and room layout.
iGibson-DPC is built on scenes from iGibson~\cite{shen2020igibson} by randomly replacing objects with different models from the same category and rendering using the iGibson simulator to render panoramas.
The selection and placement of objects in iGibson-DPC is based on heuristic algorithms, while our \dataset is manually annotated and placed 3D models are verified in terms of match and alignment to the object masks.
Moreover, iGibson-DPC contains unrealistic object arrangements (e.g., floating TV in \cref{fig:dataset_comparison}).

\section{Experiments}

We showcase the value of \dataset on the \taskfull (\task) task~\cite{zhang2014panocontext, zhang2021deeppanocontext, dong2023panocontext}.
Given an input RGB panorama, the goal is to estimate the room layout, detect objects in 2D, estimate their 3D oriented bounding boxes and also reconstruct 3D object meshes.
Our experiments show that methods trained on \dataset data benefit from its realism and generalize better when evaluated on photorealistic images.
We also investigate the role of object support hierarchy information in improving performance.

\subsection{Task setup}

\mypara{Method.}
DeepPanoContext (DPC)~\cite{zhang2021deeppanocontext} predicts the room layout, detects objects in 3D and recovers object meshes from a panorama image using a relation-based graph convolutional network and a differentiable relation optimization procedure.
Since DPC has a publicly-available implementation, we use it to benchmark the \dataset data on the \task task.
We keep all hyperparameters unchanged except lowering the relation optimization loss weight of 3D bounding box back-projection from 10 to 1, since the ground truth 2D masks are noisy.

\mypara{Datasets.}
We train and evaluate DPC on the iGibson-DPC (IG)~\cite{shen2020igibson, zhang2021deeppanocontext, dong2023panocontext}, Structured3D (S3D)~\cite{zheng2019structured3d}, and \dataset datasets.
\citet{zhang2021deeppanocontext} render 1,500 panoramas from 15 iGibson houses composed of 500+ objects spanning 57 object categories.
We use the same data and splits for IG.
Structured3D consists of 3500 houses and around 18K photo-realistic rendered panoramas in total.
We use 14K for training and the remaining 4K for testing.
Note that Structured3D does not provide ground truth object meshes.

To prepare \dataset for this task, we generate the ground truth room layout from the 3D architecture based on the camera viewpoint and obtain 3D oriented bounding boxes (OBBs) from all objects. We use 2D object masks from the Matterport3D mesh instance segmentation.
We consider three variants of \dataset based on the input panorama:
\datareal where we use the Matterport3D panoramas, 
\datasyn where we use rendered panoramas (at the same camera poses) from the annotated synthetic scenes, 
and \datamix where we combine the two types of panoramas and double the available data.  
We follow the MP3D house split and merge the train and val sets to obtain a disjoint split of 15 train and 5 test houses.
Based on the split, we have 696 annotated panoramas for train and 146 for test.
To fairly evaluate methods trained on different datasets, we curate a list of 25 object classes common to all datasets.


\begin{table}[t]
\centering
\resizebox{0.8\linewidth}{!}
{
\begin{tabular}{@{}l ccc@{}}
\toprule
Train & 2D IoU $\uparrow$ & 3D IoU $\uparrow$ & dRMSE $\downarrow$ \\
\midrule
DPC~\cite{zhang2021deeppanocontext} & 53.4 & 50.3 & 0.682  \\
\dataset-real & 55.1 & 53.1 & 0.610 \\
\dataset-syn & 59.0 & 56.1 & 0.629 \\
\dataset-mix & \textbf{59.6} & \textbf{57.0} & \textbf{0.572} \\
\bottomrule
\end{tabular}
}
\caption{
\textbf{Room layout estimation on \datareal test set.}
DPC~\cite{zhang2021deeppanocontext} was pretrained on IG and S3D.
For the last three rows, we fine-tune the pretrained weights on variants of \dataset.}
\label{tab:layout_est}
\end{table}
\begin{table}
\centering
\resizebox{\linewidth}{!}
{
\begin{tabular}{@{}ll rr rr rrrr@{}}
\toprule
\multirow{2}{*}{\rotatebox[origin=c]{90}{Test}} & \multirow{2}{*}{Train} & \multicolumn{2}{c}{3D detection $\uparrow$} & \multicolumn{2}{c}{Collision $\downarrow$} & \multicolumn{4}{c}{Attachment F1 $\uparrow$} \\
\cmidrule(lr){3-4}
\cmidrule(lr){5-6} \cmidrule(lr){7-10}
& & IoU & mAP & mesh & arch & obj & wall & floor & ceil \\
\midrule
\multirow{5}{*}{\rotatebox[origin=c]{90}{IG}} & IG & \textbf{27.5} & \textbf{30.3} & 1.662 & 2.594 & 53.1 & \textbf{76.8} & \textbf{95.0} & \textbf{86.2} \\ 
& IG+\dataset & 24.0 & 30.2 & 1.404 & 2.254 & \textbf{59.7} & 64.1 & 94.6 & 2.7 \\ 
& \dataset-real & 17.3 & 13.4 & \textbf{0.242} & 1.456 & 38.8 & 64.0 & 92.8 & 0.0 \\ 
& \dataset-syn & 23.2 & 14.2 & 0.480 & 1.938 & 48.5 & 46.7 & 93.8 & 28.6 \\ 
& \dataset-mix & 21.6 & 15.6 & 0.434 & \textbf{1.248} & 43.1 & 67.2 & 90.1 & 9.8 \\ 

\midrule
\multirow{5}{*}{\rotatebox[origin=c]{90}{S3D}} & IG & 19.5 & 3.5 & 1.016 & 2.651 & \textbf{50.9} & \textbf{68.7} & 90.8 & \textbf{11.6} \\ 
& IG+\dataset & \best{19.7} & 7.0 & 0.868 & 2.089 & 52.0 & 67.4 & \textbf{91.2} & 1.8 \\ 
& \dataset-real & 18.4 & 7.1 & 0.600 & 2.598 & 45.0 & 61.6 & 89.7 & 0.7 \\ 
& \dataset-syn & 19.0 & 4.8 & 0.644 & 2.561 & 49.3 & 49.7 & \textbf{91.2} & 2.4 \\ 
& \dataset-mix & 19.6 & \best{7.5} & \textbf{0.463} & \textbf{1.673} & 47.8 & 64.1 & 87.2 & 0.9 \\ 

\midrule
\rowcolor{offwhite}
 & IG & 15.6 & 5.9 & 0.575 & 1.959 & \textbf{53.8} & 50.7 & 51.2 & 0.0 \\
\rowcolor{offwhite}
& IG+\dataset & 17.5 & 14.1 & 0.281 & 1.267 & 49.5 & \textbf{61.6} & 58.6 & 0.0 \\
\rowcolor{offwhite}
& \dataset-real & 16.4 & 15.0 & 0.226 & 1.562 & 44.0 & 57.3 & 58.9 & 0.0 \\
\rowcolor{offwhite}
& \dataset-syn & 14.0 & 8.4 & 0.390 & 1.664 & 54.1 & 40.6 & 49.1 & 0.0 \\
\rowcolor{offwhite}
\multirow{-5}{*}{\rotatebox[origin=c]{90}{\textbf{\dataset}}} & \dataset-mix & \textbf{17.6} & \textbf{15.8} & \textbf{0.171} & \textbf{1.007} & 48.5 & 58.3 & \textbf{60.1} & 0.0\\

\bottomrule
\end{tabular}
}
\caption{
\textbf{Cross-dataset evaluation for the \taskfull task.}
We evaluate 3D detections with class-agnostic IoU and mAP at IoU of 0.15, and report object collisions. The \colorbox{offwhite}{highlighted} rows indicate the most challenging scenario.
}
\label{tab:psu_cross_train_test}
\end{table}
\begin{figure*}
\centering
\includegraphics[width=\linewidth]{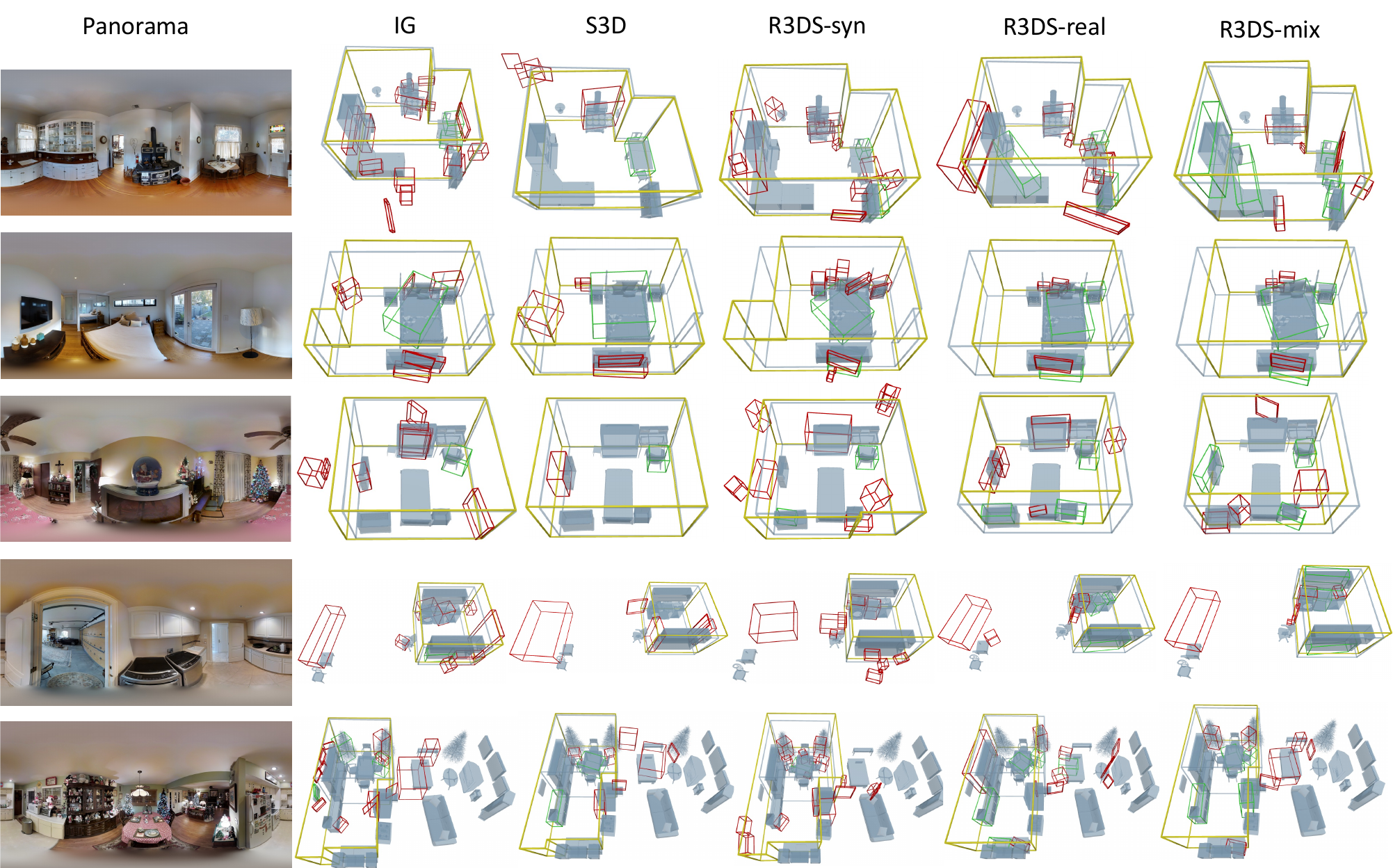}
\caption{\textbf{Qualitative results for cross-dataset \taskfull task}.
Correct and incorrect object detections shown in green and red boxes.
Ground truth room layout and meshes shown in gray color, while layout prediction is in yellow.
Training on \dataset leads to fewer errors compared to training on other datasets, especially when mixed with real data.
}
\label{fig:psu_qualitaive}
\end{figure*}

\mypara{Metrics.}
Following \citet{zhang2021deeppanocontext} we use separate metrics for room layout estimation, 3D object detection, and scene relation prediction.
For room layout estimation, we use 2D IoU for predicted 2D floorplan, 3D IoU for lifted 3D room geometry, and dRMSE for predicted depth with respect to the camera location.
For 3D object detection, we report bounding box-based class-agnostic 3D IoU as well as mean average precision (mAP) across the 25 object classes, where an IoU greater than 0.15 counts as a ``true'' result.
For scene relation prediction, we report F1 scores for relation classification.
We also report the average number of objects colliding with each other or with architectural structures (wall, floor, ceiling).
Specifically, we follow \citet{zhang2021deeppanocontext} and measure collisions using the Separating Axis Theorem (SAT) to test whether object bounding boxes overlap.
Since bounding box-based collision is a poor proxy for real-world physical collision, we also compute mesh-based collision by checking if the meshes for object pairs have any interpenetrating triangles~\cite{Karras:2012:MPC:2383795.2383801,Tzionas:IJCV:2016}.

\subsection{Results}

\mypara{1) Does \dataset help DPC generalize to real images?}
Since the original DPC work only trained and evaluated on synthetic data, it is unclear how well it performs on realistic panoramic imagery.
We hypothesize that training on \dataset will lead to better performance.
We separately show results on room layout estimation (\cref{tab:layout_est}) and 3D object detection.

\emph{Room layout estimation.}
For room layout estimation, DPC uses HorizonNet~\cite{sun2019horizonnet} pretrained on iGibson (IG) and Structured3D (S3D) panoramas.
This model achieves good performance on IG data (91.0 3D IoU).
When directly testing the official pretrained model on \datareal panoramas, we notice a significant performance drop compared to results on the rendered panoramas from iGibson (\cref{tab:layout_est} shows that the 3D IoU drops to 50.3).
By finetuning the pretrained model with \datareal, we can predict more precise room layouts for real cluttered scenes.
Even only trained on \datasyn, we outperform the original DPC model by 5.6\% and 5.8\% on 2D and 3D IoU, respectively.
This is likely due to renderings from \datasyn reflecting more realistic object arrangements in a room instead of pushing all objects against walls.
Best performance on 2D IoU (59.6), 3D IoU (57.0) and depth RMSE (0.572) is achieved by fine tuning on \datamix.

\emph{Object detection.}
For 3D object detection, we train DPC on different data settings and conduct a cross-dataset evaluation (see \cref{tab:psu_cross_train_test}).
To investigate how models perform on out-of-distribution scenes, we evaluate models on Structured3D, as its images are near-realistic.
To explore whether DPC training benefits from \dataset given the same amount of data, we create a special data input \IGplusours that combines iGibson and \dataset panoramas by randomly replacing half (500) of iGibson data with \datareal data.
The results show that \IGplusours performs almost the same as \IG with fewer collisions on iGibson, but it remarkably outperforms \IG on the test set of \emph{\dataset} and \STD by 8.2 and 3.5 improvements on 3D mAP, respectively.
It also averages 0.221 fewer mesh collisions.
There are noticeable performance gaps on iGibson for models trained on \dataset data likely due to the data domain shift.
Among the three variants of \dataset data, \datamix outperforms the others on all three test sets regarding 3D IoU and 3D mAP with the fewest mesh and architecture collisions.
Although \datasyn underperforms on \emph{\dataset} and \STD test sets, it achieves better performance than \IG with even less data.

\emph{Scene relation classification.}
We report F1 scores for identifying attachment relationships of objects to other objects and architecture elements (see \cref{tab:psu_cross_train_test}).
We note that models trained with synthetic renderings perform better than those trained on real images.
That is because synthetic renderings present cleaner and simpler scenes with fewer objects than real world and simpler illumination such that DPC finds it easier to learn object-object and object-architecture relations.
Also, note that the predictions of object-ceiling attachments can be extremely low because few objects are attached to the ceiling in the ground truth data.
We show qualitative examples in \Cref{fig:psu_qualitaive}.

\begin{table}
\centering
\resizebox{\linewidth}{!}
{
\begin{tabular}{@{}l rr rr rrr@{}}
\toprule
 &  \multicolumn{2}{c}{3D detection } & \multicolumn{2}{c}{Collision $\downarrow$}  & \multicolumn{3}{c}{Support F1 $\uparrow$} \\
\cmidrule(lr){2-3} \cmidrule(lr){4-5} \cmidrule(lr){6-8}
Train & IoU$\uparrow$ & mAP$\uparrow$ & mesh & arch & obj & floor & ceil \\
\midrule
IG & 14.2 & 5.2 & 0.703 & 1.639 & \best{4.1} & 85.3 & 0.0 \\
S3D & 16.4 & \best{10.0} & \best{0.112} & 1.226 & 3.1 & \best{86.8} & 0.0 \\
IG+S3D & \best{17.1} & 9.7 & 0.133 & \best{1.162} & 3.3 & 84.8 & 0.0 \\
\bottomrule
\end{tabular}
}
\caption{Performance of models trained on three synthetic datasets (IG, S3D, and IG+S3D) evaluated on the \datafull dataset, where ``full'' indicates all 840 panoramas are used for testing.}
\label{tab:psu_full_set}
\end{table}
\begin{figure}
\centering
\includegraphics[width=\linewidth]{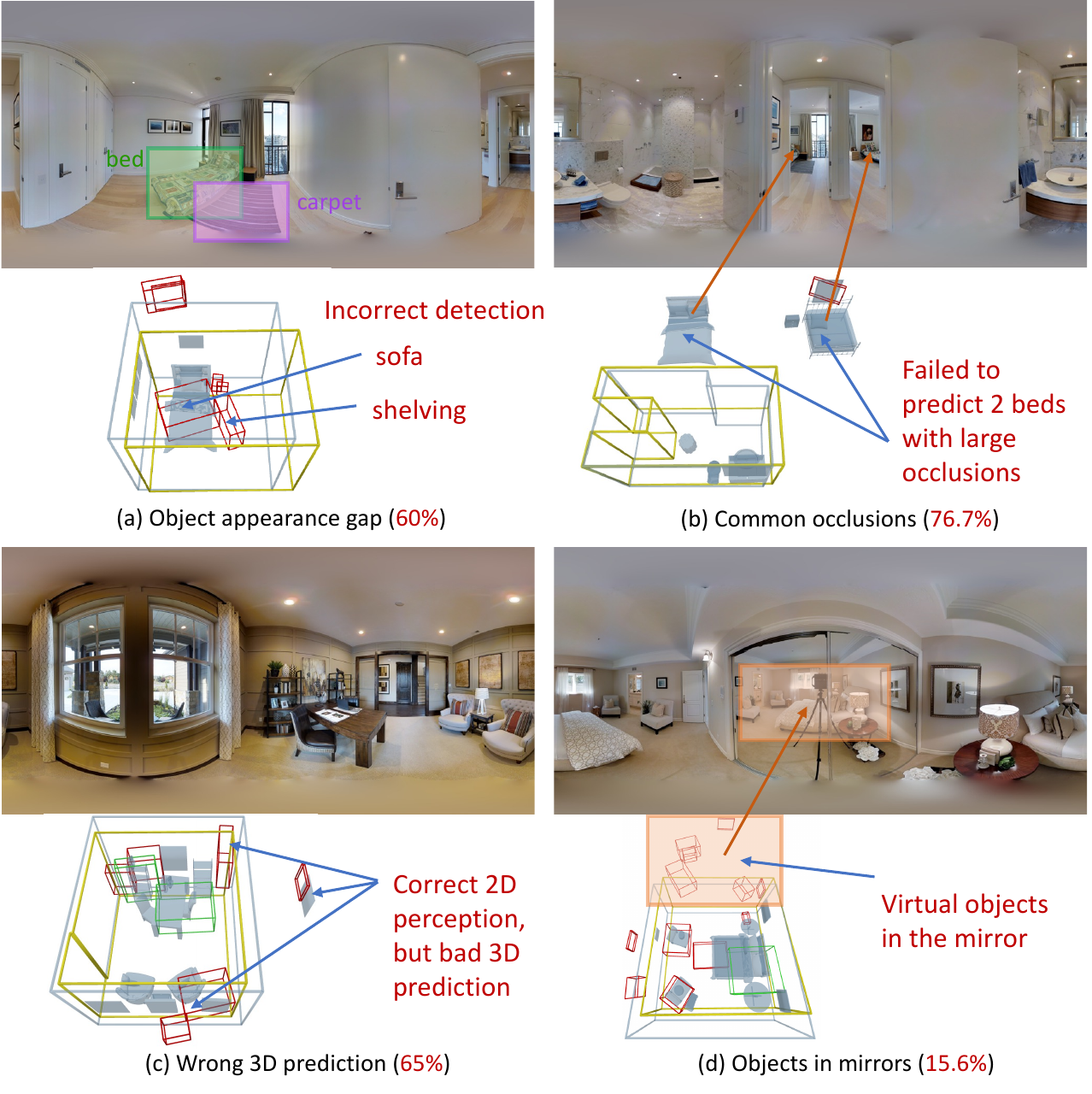}
\caption{
\textbf{Error analysis on \dataset panoramas using DPC pretrained on S3D.}
Typical prediction errors are categorized into 4 groups:
(a) Appearance gap between real and synthetic images.
(b) Detection failure due to common occlusions.
(c) Bad performance of monocular 3D prediction.
(d) Perception error for objects in mirrors.
Numbers in parentheses mean the error rates.
}
\label{fig:error_analysis}
\end{figure}

\mypara{2) Is \dataset a challenging, high-quality test set?}
How would a model trained on pure synthetic data perform on complex real data (\dataset)?
Due to its modest scale, we propose using \dataset as a challenging, high-quality test set rather than a train set.
Specifically, we evaluate the synthetic-to-real performance of DPC by training on iGibson and/or Structured3D and testing on all panoramas in \dataset-real.
\Cref{tab:psu_full_set} shows that a model trained with Structured3D performs the best (10.0 3D mAP and 0.112 mesh collision) as it observes the most photo-realistic images.
DPC benefits from the synthetic data for higher bounding box IoUs, since it possesses accurately aligned 3D bounding box and more unoccluded objects. However, mAP performance is lower due to worse object recognition ability.
All models struggle to predict correct object-wise support relations but do a better job of predicting object-floor support relations. 

We conduct error analysis on 120 randomly sampled panoramas using the model pretrained on \STD to identify typical errors (see \cref{fig:error_analysis}).
Errors are categorized into 4 groups:
(a) 60\% panoramas have 2D perception errors due to the synthetic-to-real appearance gap;
(b) 76.7\% panoramas show detection failures due to occlusions;
(c) 65\% panoramas exhibit correct 2D detections but fail to correctly perform 3D predictions; and
(d) 15.6\% out of 45 panoramas with mirrors mistakenly predict virtual objects in mirrors.

\begin{figure}
\centering
\includegraphics[width=\linewidth]{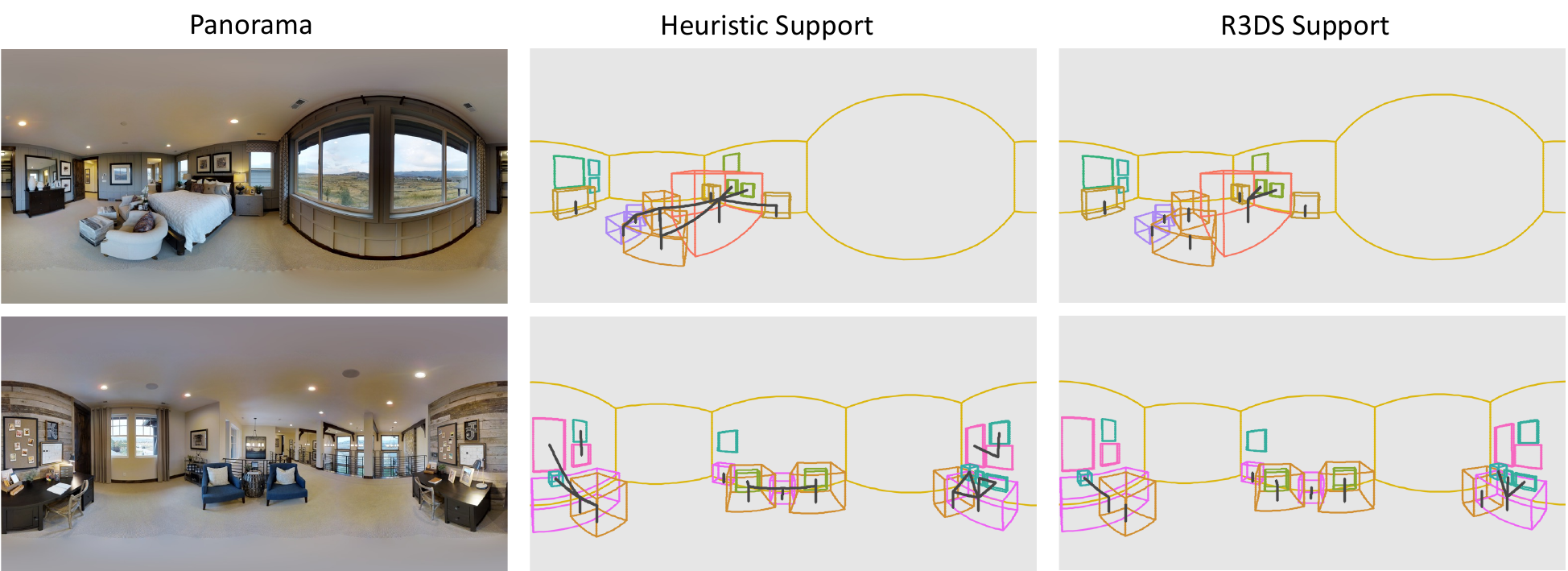}
\caption{\textbf{Comparison between heuristic and \dataset support relations.} Dark lines between objects show support relations.
Heuristic support often mistakenly assigns relations to spatially close objects.
In contrast, \dataset support relations are annotated.
}
\label{fig:supp_rel_comparison}
\end{figure}
\begin{table}
\centering
\resizebox{\linewidth}{!}
{
\begin{tabular}{@{}l lrr rr rrr@{}}
\toprule
& & \multicolumn{2}{c}{3D detection} & \multicolumn{2}{c}{Collision $\downarrow$} & \multicolumn{3}{c}{Support F1 $\uparrow$} \\
\cmidrule(lr){3-4} \cmidrule(lr){5-6} \cmidrule(lr){7-9}
Train & Supp. & IoU$\uparrow$ & mAP$\uparrow$ & mesh & arch & obj & floor & ceil \\
\midrule
\multirow{3}{*}{\dataset-real} & none & 16.4 & 15.0 & 0.226 & 1.562 & - & - & -  \\
& heur & 16.2 & 14.1 & \best{0.219} & \best{1.329} & 3.2 & 69.9 & 0.0 \\
& anno & \best{16.6} & \best{15.9} & 0.349 & 1.404 & \best{38.5} & \best{94.4} & 0.0 \\
\midrule
\multirow{3}{*}{\dataset-syn} & none & 14.0 & \best{8.4} & 0.390 & 1.664 & - & - & -  \\
& heur & \best{14.6} & 7.8 & \best{0.281} & 1.301 & 4.6 & 82.9 & \best{52.6} \\
& anno & 14.3 & 8.2 & 0.349 & \best{1.219} & \best{32.0} & \best{95.0} & 0.0  \\
\midrule
\multirow{3}{*}{\dataset-mix} & none & 17.6 & 15.8 & 0.171 & \best{1.007} & - & - & -  \\
& heur & \best{19.2} & 17.7 & \best{0.151} & 1.267 & 3.6 & 83.7 & 0.0 \\
& anno & 18.6 & \best{18.2} & 0.158 & 1.308 & \best{12.0} & \best{96.2} & \best{85.8} \\
\bottomrule
\end{tabular}
}
\caption{Performance on \datareal of DPC models trained on variants of \dataset with different support relation settings.  We compare the original model without support (\emph{none}) against models supervised with support that is heuristically computed (\emph{heur}) or annotated from \dataset (\emph{anno}). Classification results are evaluated on annotated ground-truth scene hierarchy.}
\label{tab:psu_support_rel}
\end{table}

\mypara{3) Are \dataset support relations helpful for \task?}
We investigate whether the support relationships between objects provided in our \dataset scene hierarchy help boost performance of holistic scene understanding. We augment DPC's Relation Scene-GCN module with additional support relation prediction branches. 
Besides obtaining explicitly annotated scene support relations from \dataset, it is also possible to compute heuristic support relations from object bounding boxes.
Specifically, an object is supported by another if their bounding boxes intersect within tolerance distance of 0.1m and the centroid of the former object is higher than that of the latter.
Support by wall/floor/ceiling is calculated in the same way without the height judgment.
This definition is similar to how DPC defines object attachment.
\Cref{fig:supp_rel_comparison} compares these two ways of computing support relations, showing that heuristic computation can mistakenly designate support relations to two nearby objects.
\Cref{tab:psu_support_rel} shows that incorporating support relation prediction indeed influences the performance of DPC.
Heuristic support information may worsen 3D object detection (mAP in \datareal and \datasyn), but it eliminates mesh collisions the most.
Learning support relations from \dataset annotations leads to a 2.4 improvement on mAP in \datamix, although the classification F1 score is low.

\begin{table}
\centering
\resizebox{\linewidth}{!}
{
\begin{tabular}{@{}l c cccc@{}}
\toprule
\multirow{2}{*}{Datasets} & \multirow{2}{*}{Mesh Collisions} & \multicolumn{4}{c}{Box Collisions}  \\
\cmidrule(lr){3-6}
& & obj-obj & obj-wall & obj-floor & obj-ceil \\
\midrule
IG & - & 1.185 & 0.075 & 0.000 & 0.790  \\
\dataset & 0.0006 & 2.823 & 0.388 & 0.035 & 0.064 \\
\bottomrule
\end{tabular}
}
\caption{
\textbf{Comparison of the average number of bounding box and mesh-based object collisions per scene in IG and \dataset.}
\dataset exhibits more bounding box-based collisions, but almost none of these are actual physical collisions between object meshes.
Measuring collisions between bounding boxes is a poor collision measure for fully-populated, real-world scenes.
}
\label{tab:dataset_collisions}
\end{table}

\begin{table}
\centering
\resizebox{\linewidth}{!}
{
\begin{tabular}{@{}l cc cc@{}}
\toprule
\multirow{2}{*}{Rel. Opt.} & \multirow{2}{*}{3D mAP $\uparrow$} & \multirow{2}{*}{Mesh Collisions $\downarrow$} & \multicolumn{2}{c}{Box Collisions $\downarrow$}  \\
\cmidrule(lr){4-5}
& & & obj-obj & obj-arch \\
\midrule
DPC & 18.2 & 0.158 & 0.062 & 1.308   \\
w/o obj col & 18.9 & 1.342 & 1.130 & 1.301  \\
w/o obj col+tch & 19.6 & 1.219 & 1.062 & 1.295  \\
w/ mesh col & 19.7 & 1.027 & 0.856 & 1.394  \\
\bottomrule
\end{tabular}
}
\caption{
\textbf{Ablation of relation optimization (RO) on \dataset-real.}
The 2nd and 3rd row remove optimization terms in RO.
The last row replaces bounding-box collisions with mesh collisions.
}
\label{tab:ablation_ro}
\end{table}

\mypara{4) Is relation optimization (RO) effective on \dataset?}
Test-time relation optimization (RO) was introduced by \citet{zhang2021deeppanocontext} to reduce physical violations, floating objects, and misalignment between objects and architecture.
The original cost function based on bounding box collisions succeeds in optimizing object poses, since there are few such collisions in the IG data originally used for evaluation (see in \cref{tab:dataset_collisions}). However, the same data assumption does not hold for \dataset, which has more bounding-box-based collisions but nearly zero mesh-based collisions.
We ablate the design of RO on \dataset-real in \Cref{tab:ablation_ro}.
By removing two optimization terms (bounding-box-based object-wise collision and touching step-by-step), the model outperforms the original one in 3D mAP (+1.4) but degrades in mesh-based and box-based collisions.
We show that using mesh-based collision optimization leads to the best performance.
The increase in collisions is unsurprising as the \dataset data reflects more cluttered real interiors.

\mypara{Limitations.}
Our dataset construction relied on 3D architectures for each Matterport3D scan which are simplifications of the geometry of the real environment.
One issue is imperfect wall positions, resulting in objects attached to these virtual walls being offset from the true surface.
In addition, objects in our 3D scenes were placed without regard to the materials, meaning that the detailed surface appearance does not match that of the observed object.
Future work can investigate transfer of surface appearance to the synthetic objects by projecting textures from the RGB-D data and 3D reconstructed meshes.

\section{Conclusion}

We introduced the \dataset dataset.
\dataset provides more complete, densely populated, and richly annotated synthetic 3D scene proxies of real-world environments with linked panoramic images.
We showed the usefulness of \dataset on the \taskfull task.
Our experiments demonstrate the value of realistic synthetic recreations in this task, in particular through the use of object support information.
While we focused on the \task task, \dataset can also be useful for other tasks such as single-view shape retrieval, single-view object pose estimation, and panoramic scene graph prediction.

\mypara{Acknowledgements.}
This work was funded in part by a CIFAR AI Chair, a Canada Research Chair, NSERC Discovery Grant, NSF award \#2016532, and enabled by support from \href{https://www.westgrid.ca/}{WestGrid} and \href{https://www.computecanada.ca/}{Compute Canada}. Daniel Ritchie is an advisor to Geopipe and owns equity in the company. Geopipe is a start-up that is developing 3D technology to build immersive virtual copies of the real world with applications in various fields, including games and architecture. We thank Madhawa Vidanapathirana, Weijie Lin, and David Han for help with development of the annotation tool, and Denys Iliash, Mrinal Goshalia, Brandon Robles, Paul Brown, Chloe Ye, Coco Kaleel, Elizabeth Wu and Hannah Julius for data annotation, and Ivan Tam, Austin Wang, and Ning Wang for feedback on the paper draft.

{
\small
\bibliographystyle{ieeenat_fullname}
\bibliography{main}
}


\clearpage
\appendix


In this supplement, we provide additional examples and statistics for our \dataset dataset (\cref{sec:supp-dataset}) and  details on our annotation interface (\cref{sec:supp-annotation-interface}). 

\begin{figure}[b!]
\centering
\includegraphics[width=\linewidth]{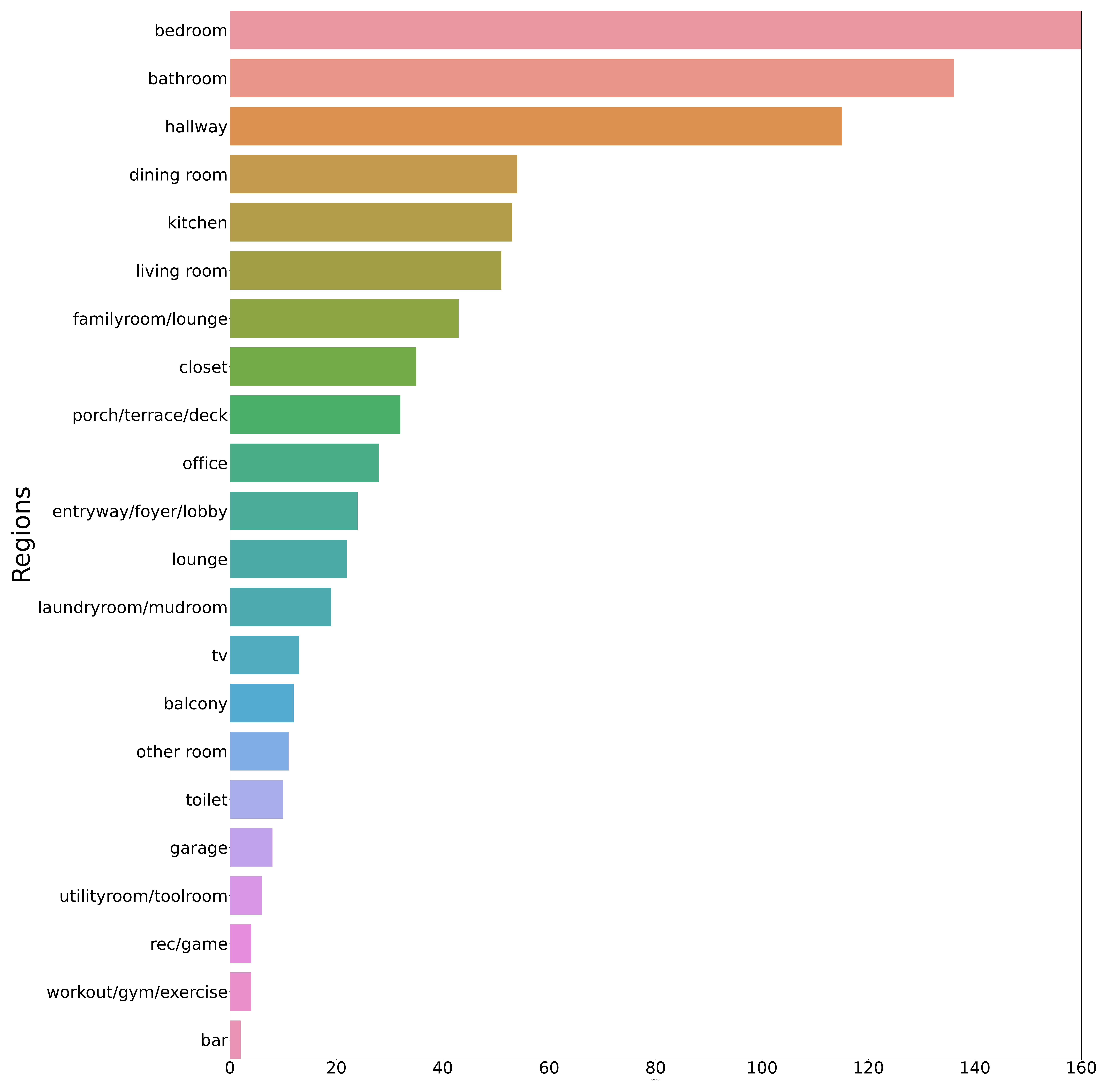}
\caption{
Histogram of all region (i.e. room) types in our \dataset dataset.
The three most common region types are bedroom, bathroom and hallway, but there is a long-tail distribution with many other region types.
}
\label{fig:region_type_plot}
\end{figure}

\begin{figure}[b!]
\centering
\includegraphics[width=\linewidth]{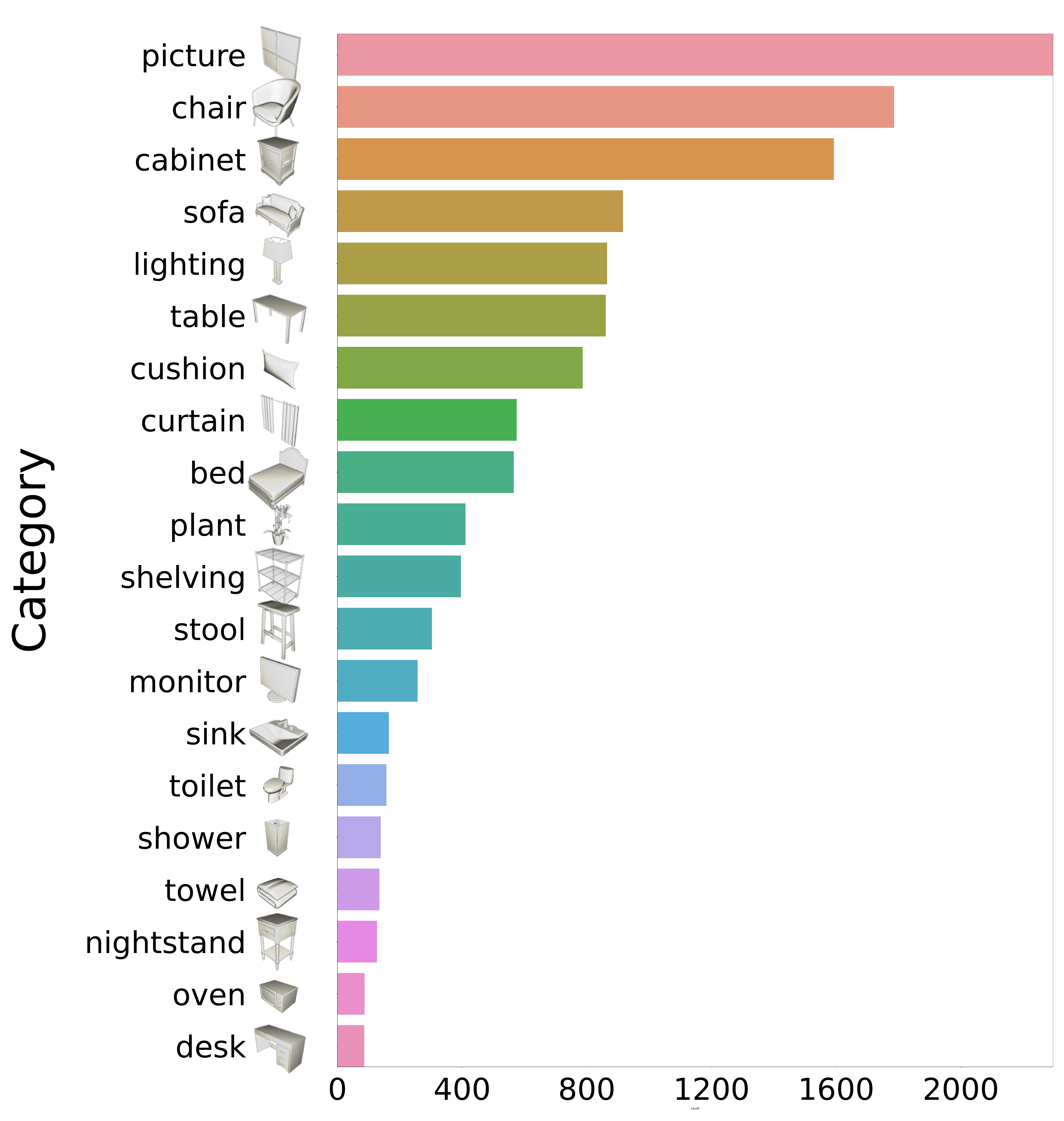}
\caption{
\textbf{Coarse Object Categories.} Histogram of the 20 most common object categories in \dataset.
We see that the scenes in our dataset exhibit a long-tail distribution over common object categories occurring in real-world scenes.
}
\label{fig:category_plot}
\end{figure}

\begin{figure}
\centering
\includegraphics[width=0.9\linewidth]{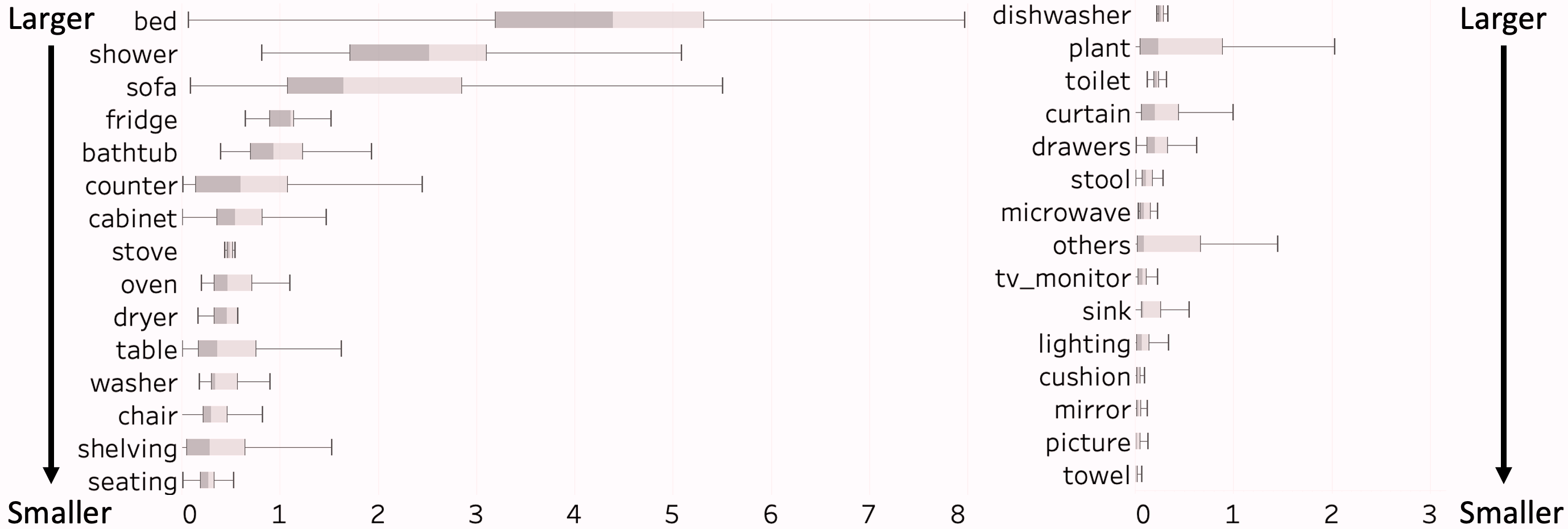}
\caption{
Box plot of the physical size distribution (measured by volume in $\text{m}^3$) per category, showing a broad spectrum of sizes with several small categories (right side).
}
\label{fig:object_sizes}
\end{figure}

\section{\dataset dataset examples and statistics}
\label{sec:supp-dataset}

\begin{figure}[t]
\centering
\includegraphics[width=\linewidth]{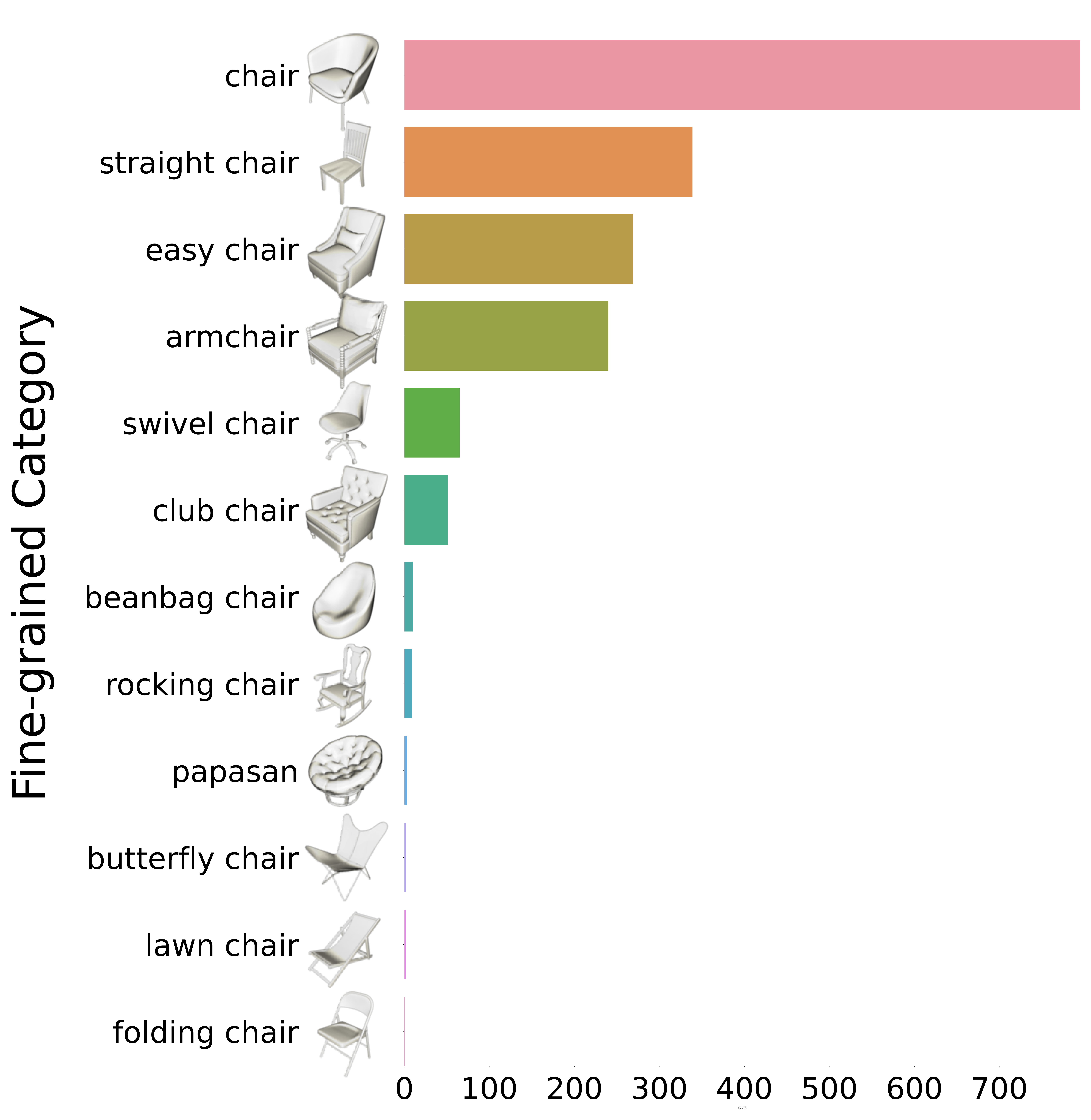}
\caption{
Histogram of the fine-grained categories for `Chair'.
We see a variety of types of chair instances present in scenes from our dataset.
}
\label{fig:category_plot_chair}
\end{figure}

\begin{figure}[t]
\centering
\includegraphics[width=\linewidth]{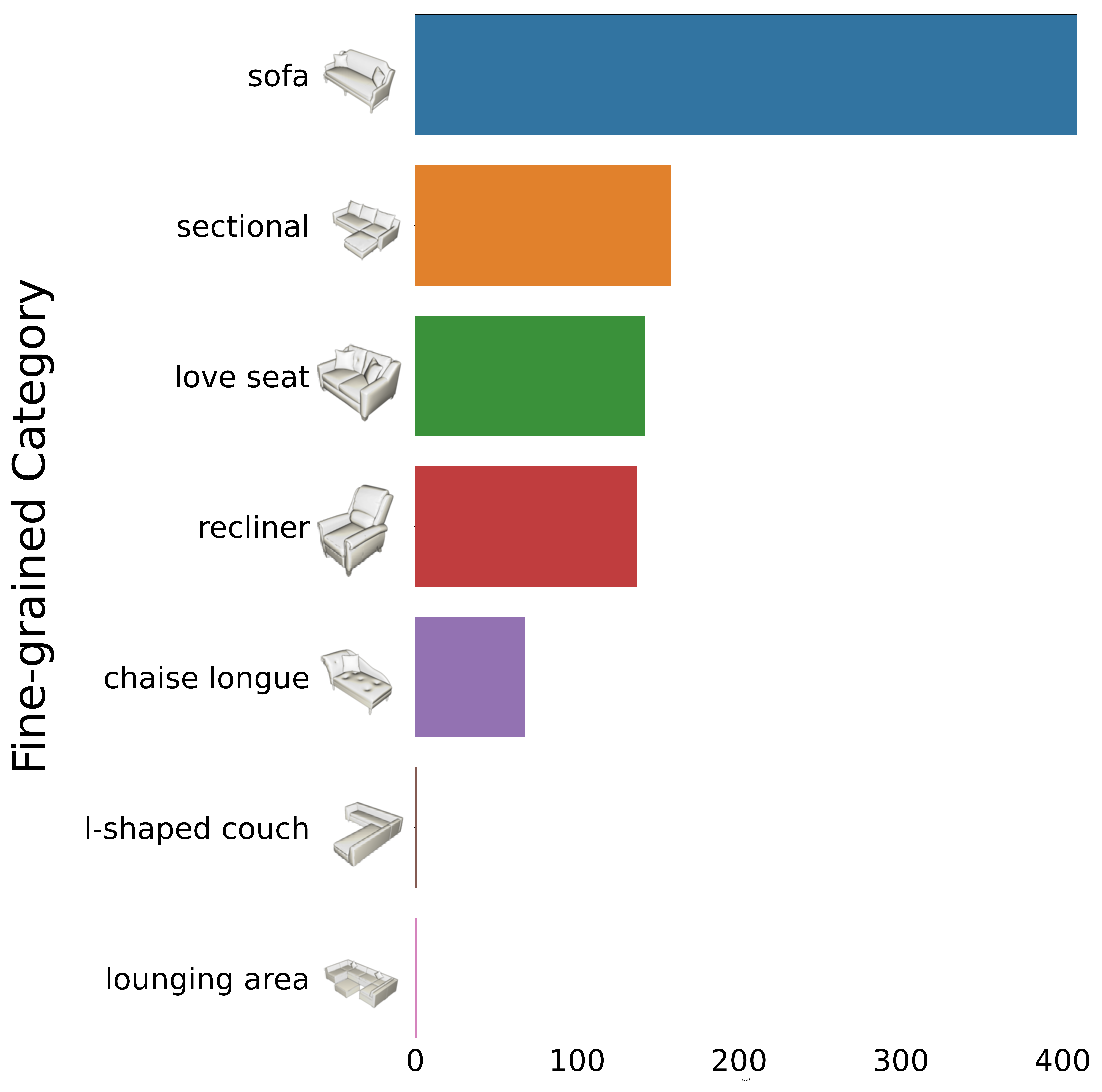}
\caption{
Histogram of the fine-grained categories for `Sofa'.
There are a variety of sofas in our scenes.
}
\label{fig:category_plot_sofa}
\end{figure}

\begin{figure}[t]
\centering
\includegraphics[width=\linewidth]{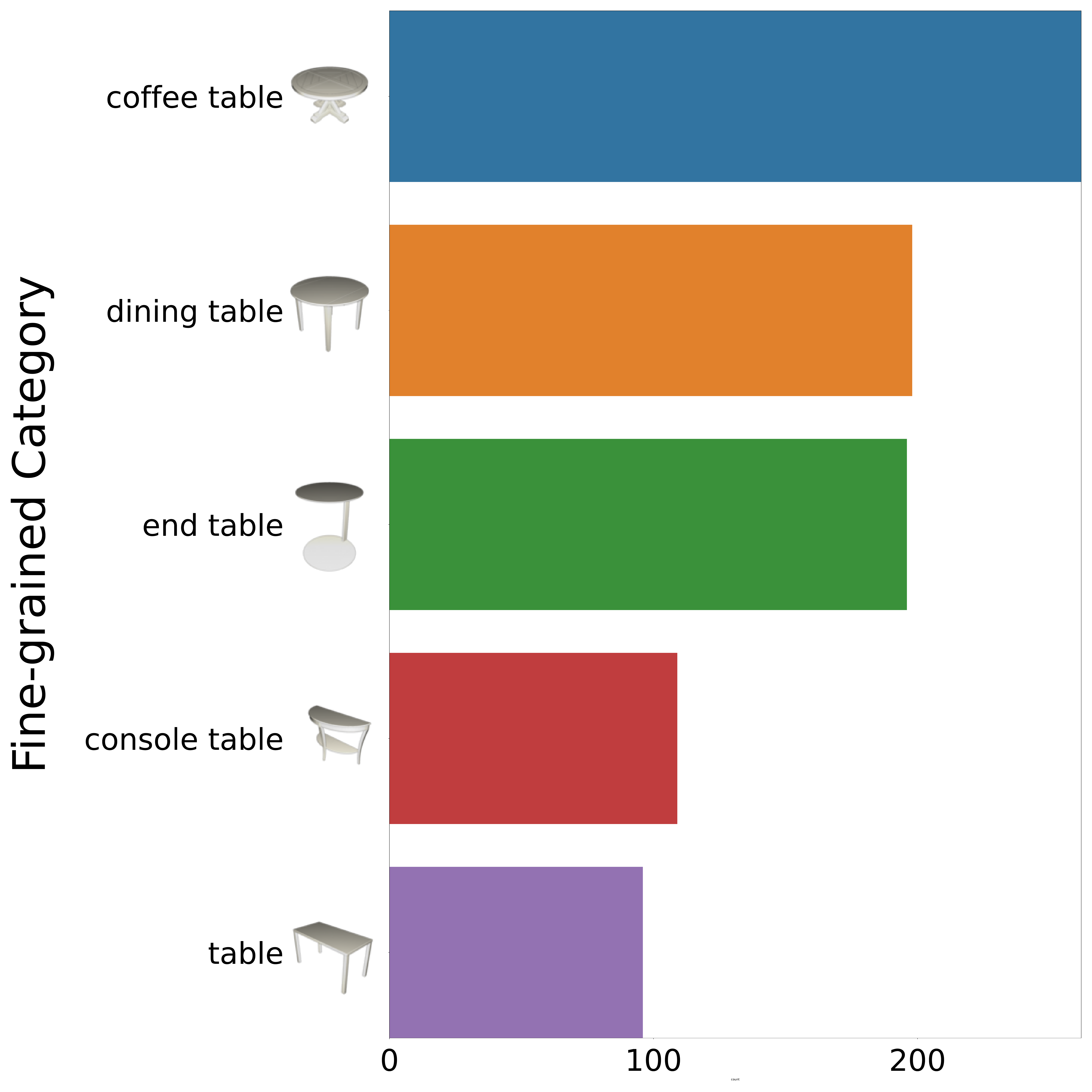}
\caption{
Histogram of the fine-grained categories for `Table'.
A variety of tables are present in our dataset scenes.
}
\label{fig:category_plot_table}
\end{figure}

\begin{figure}[t]
\centering
\includegraphics[width=\linewidth]{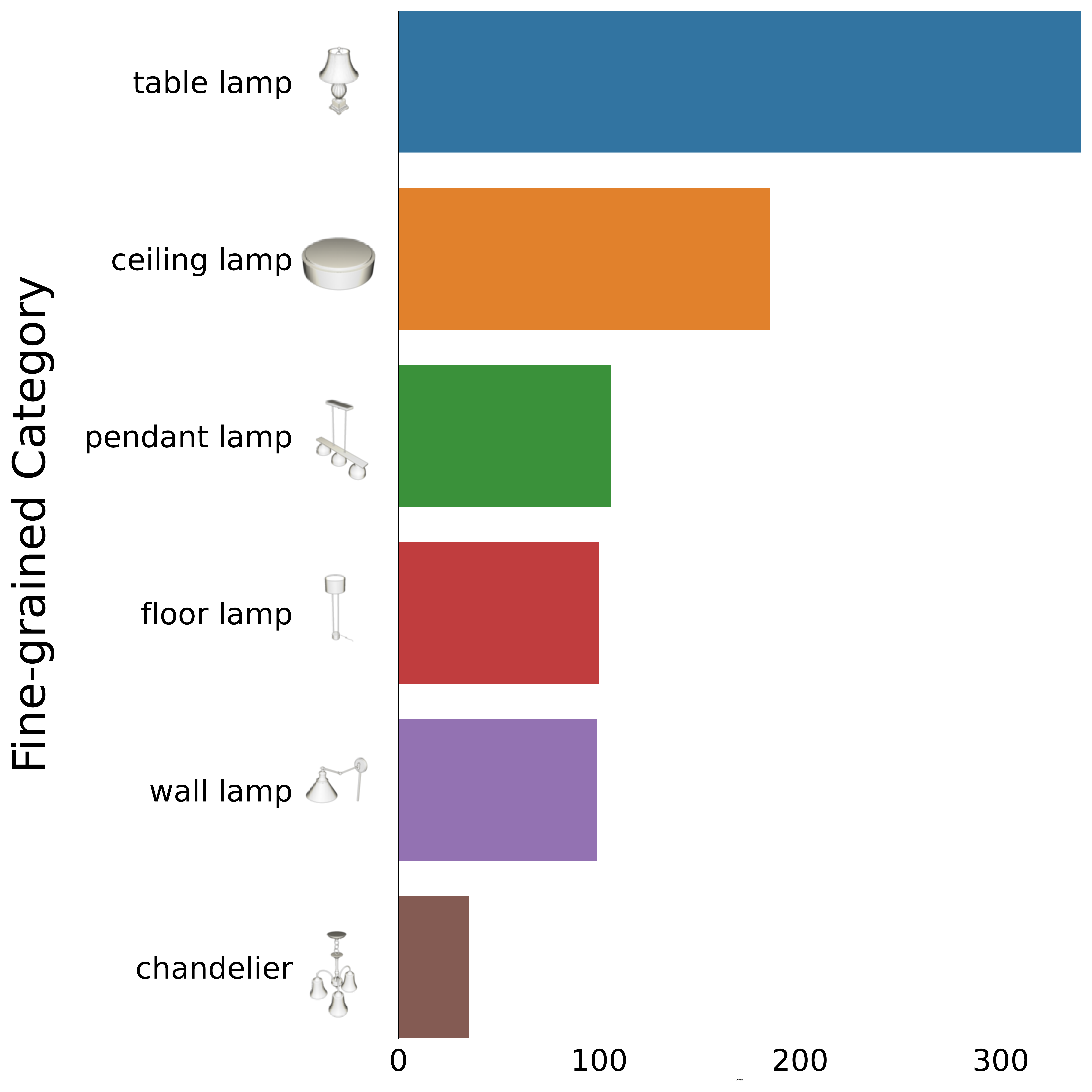}
\caption{
Histogram of the fine-grained categories for `Lighting'.
Like chairs, there are a variety of types of lighting fixtures in our scenes.
Note that these types of lights are also found in dramatically different support relations with the architecture and other objects (e.g., table lamp vs ceiling lamp vs floor lamp).
}
\label{fig:category_plot_lighting}
\end{figure}

\begin{figure*}[t]
\centering
\includegraphics[width=\linewidth]{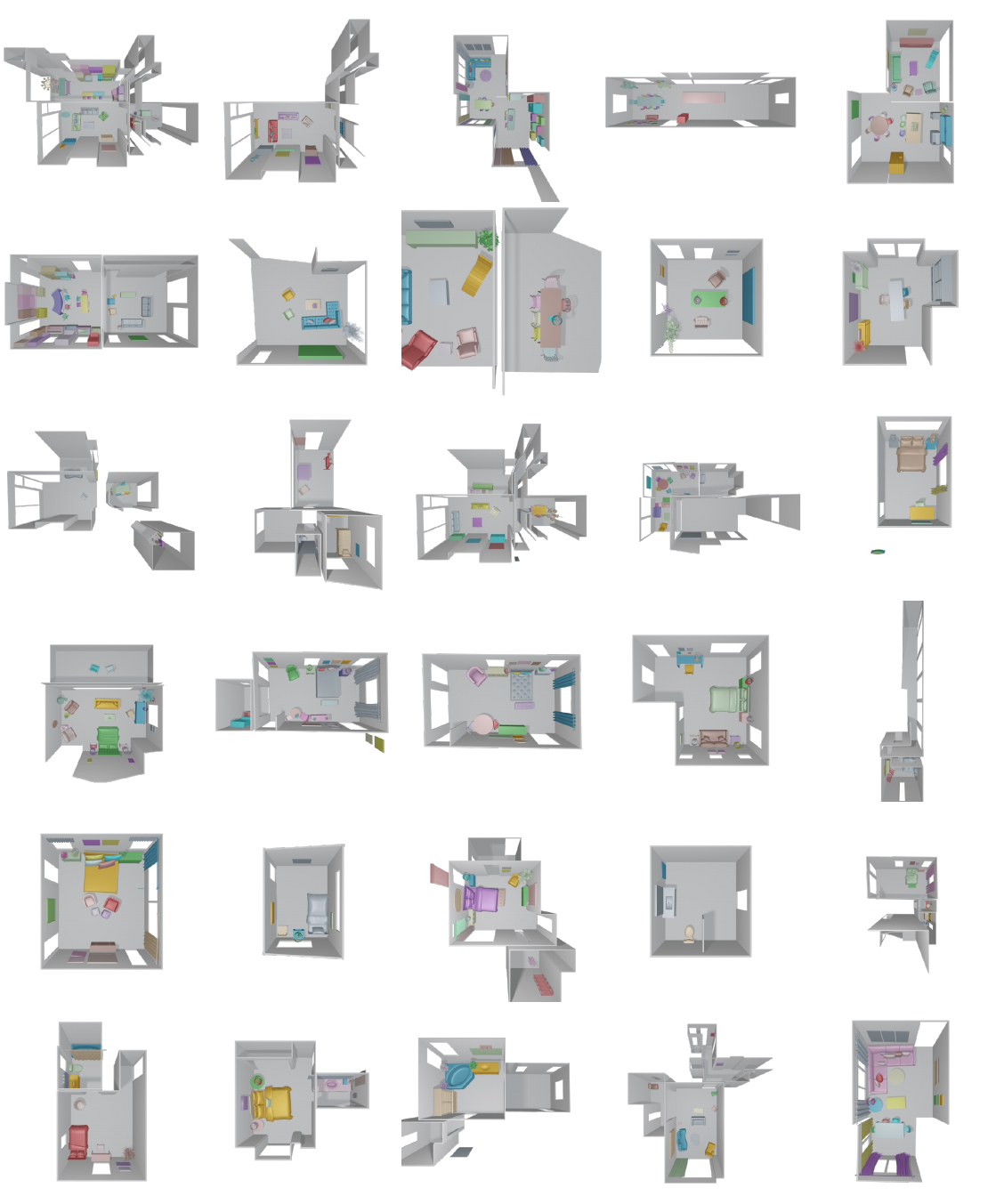}
\caption{Top-down overviews of various annotated scenes with objects colored by instances.
Our dataset covers several region types including kitchen, bedroom, bathroom, office, lounge room and more.
The object arrangements are dense, with objects supported by other objects (e.g., pillows on beds and couches) and by architectural elements (e.g., paintings on walls and curtains on windows).
}
\label{fig:annotation_topdown}
\end{figure*}

\begin{figure*}[t]
\centering
\includegraphics[width=\linewidth]{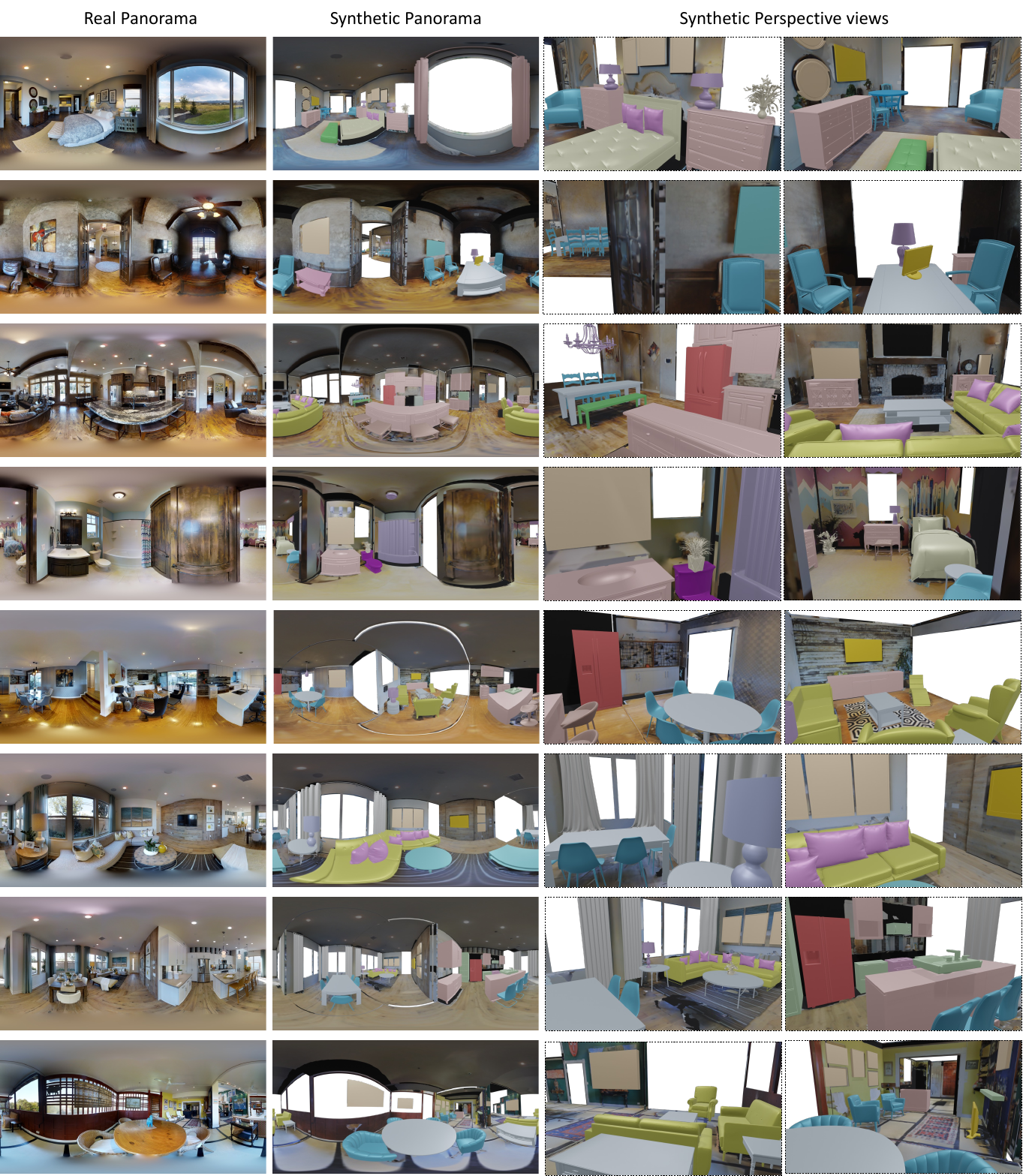}
\caption{
Examples of real-world panoramas and corresponded synthetic panoramas.
The synthetic scenes are rendered using the annotated textured 3D architecture and all placed synthetic objects, from the same camera pose as the original panorama. The rightmost two columns show additional sampled perspective views within each scene.
Objects are colored according to semantic category label.
The scenes are densely populated with objects plausibly arranging on other objects and with respect to the architecture.
}
\label{fig:annotation_example}
\end{figure*}

We show a histogram of the region types covered by our dataset (\Cref{fig:region_type_plot}), and histograms of the object categories (\Cref{fig:category_plot,fig:category_plot_chair,fig:category_plot_sofa,fig:category_plot_table,fig:category_plot_lighting}).
We first show a histogram of the 20 most commonly occurring coarse object categories in \Cref{fig:category_plot} and then fine-grained category distributions for some broader object categories such as `Chair', `Sofa', `Table' and `Lighting' in \Cref{fig:category_plot_chair,fig:category_plot_sofa,fig:category_plot_table,fig:category_plot_lighting}.
We also present a box plot of the physical size distribution (measured by volume in $\text{m}^3$) per category in \Cref{fig:object_sizes}.

We show additional qualitative examples of scenes in our \dataset dataset in \Cref{fig:annotation_topdown,fig:annotation_example}.

\begin{figure}[t]
\centering
\includegraphics[width=0.95\linewidth]{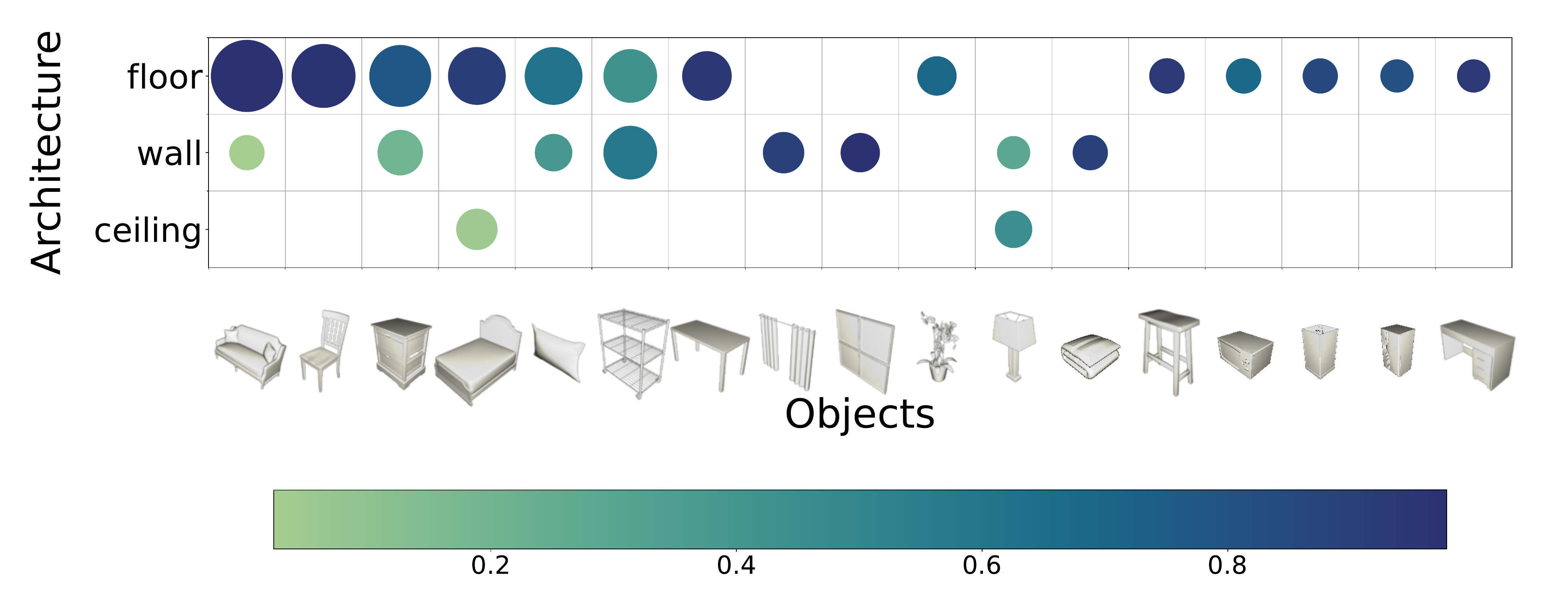}
\caption{
\textbf{Object-to-architecture support.} The plot shows support between objects and architecture elements (floor, wall, ceiling). The circle radius indicates the average number of the object type (ranges from 5 to 24) supported by a given arch type per panorama, whereas the color indicates the number of times the object type is supported by that arch type out of the total number of times the object type appears. For example, we see that on average, same number of shelves can be found on the floor and the wall in a panorama (circle radius), but overall shelves appear more on the wall than on the floor (color).
}
\label{fig:obj_to_arch_support_plot}
\end{figure}

\begin{figure}[t]
\centering
\includegraphics[width=0.95\linewidth]{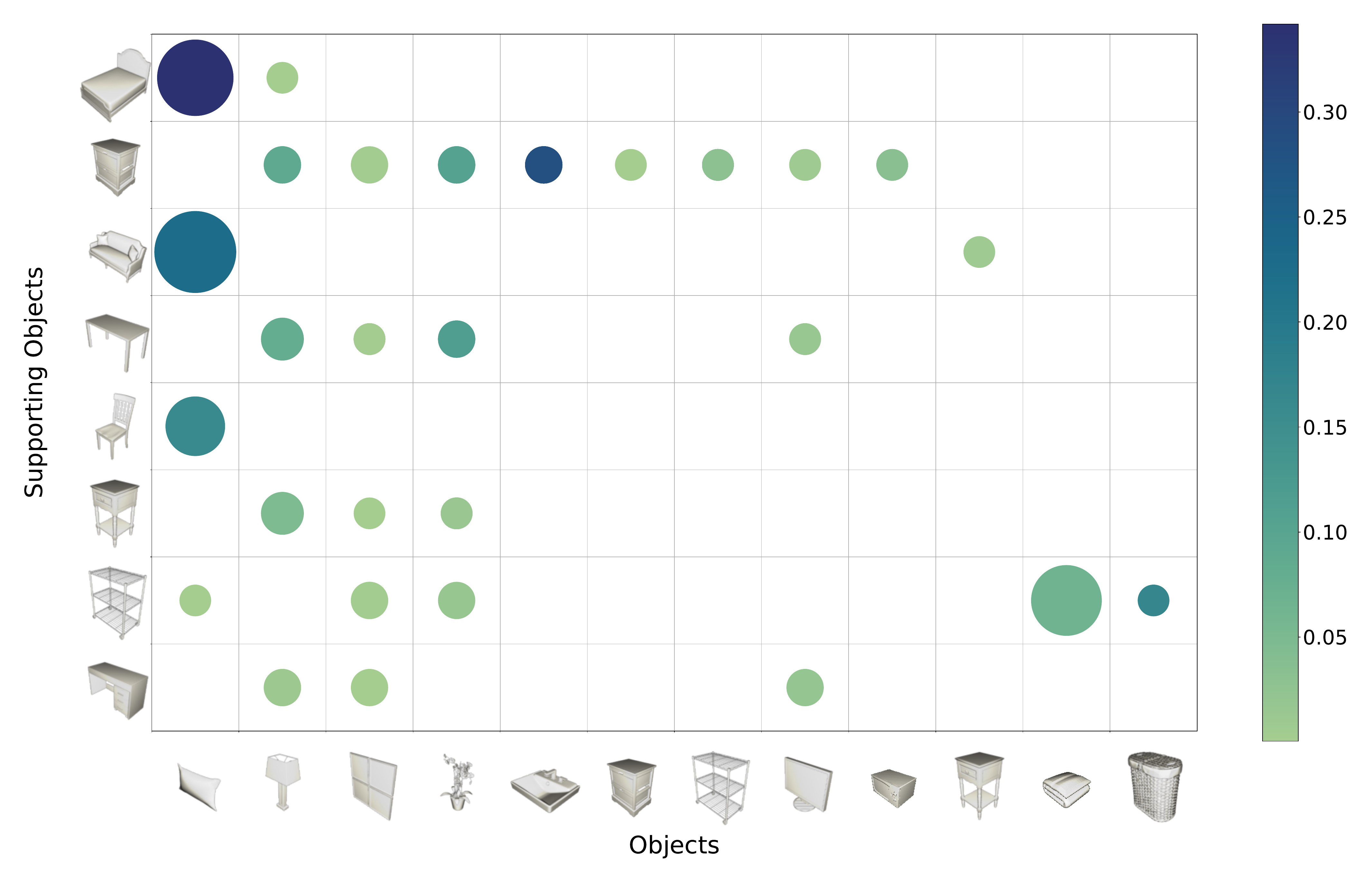}
\caption{
\textbf{Object-to-object support.} The plot shows support between various objects in \dataset. The circle radius indicates the average number of the object type (ranges from 1 to 10) supported by the parent object type per panorama, whereas the color indicates the number of times the object type is supported by the parent object type out of the total number of times the object type appears in the dataset. For example, we see that on average, more pillows are found on a couch than on a bed in a panorama (circle radius), but overall pillows appear more frequently on the bed in the dataset (color).
}
\label{fig:obj_to_obj_support_plot}
\end{figure}

\begin{figure}[t]
\centering
\includegraphics[width=\linewidth]{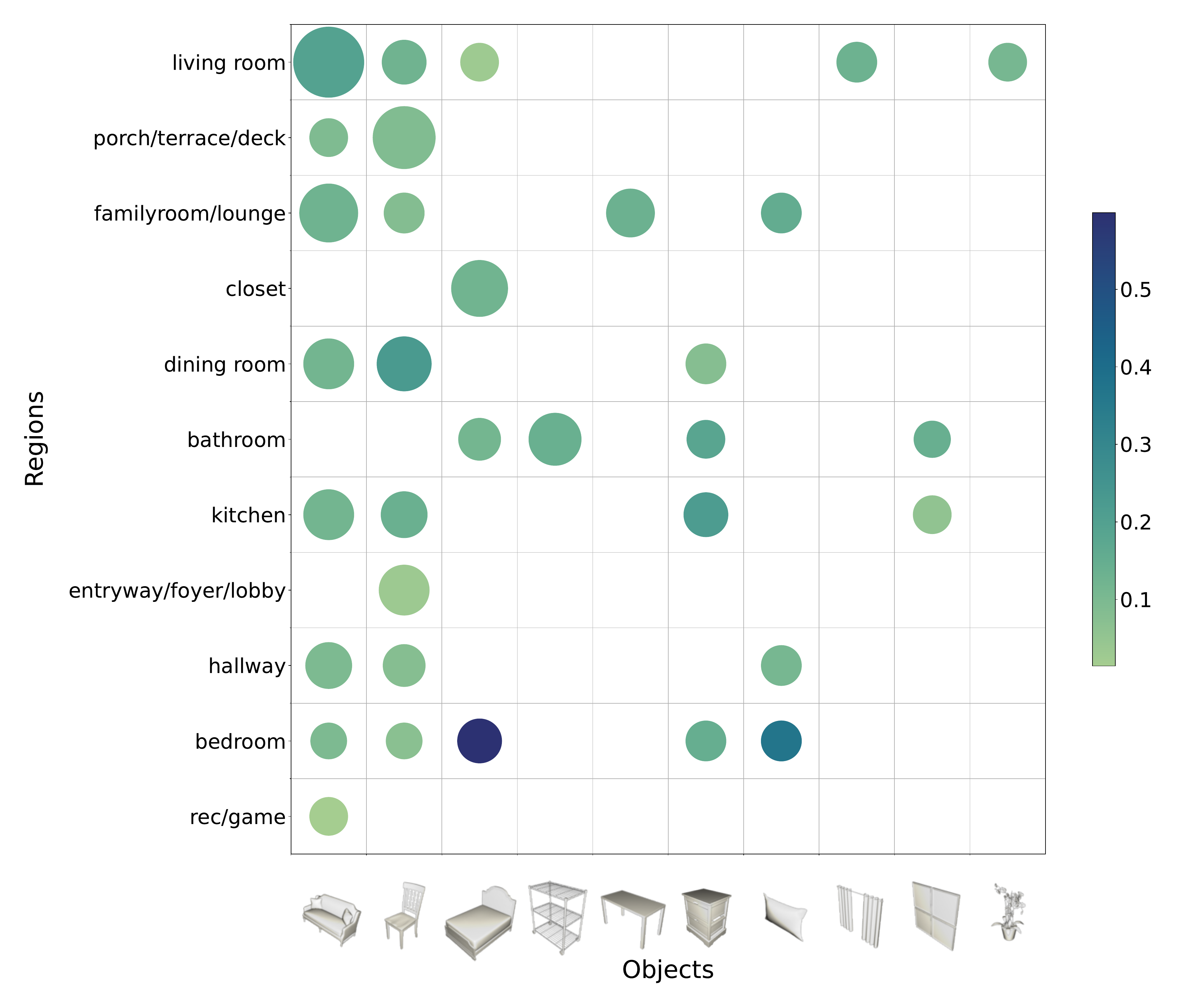}
\caption{
\textbf{Object-to-region.} The plot shows objects found in various regions in \dataset. The circle radius indicates the average number of the object type (ranges from 7 to 24) found in a region per panorama, whereas the color indicates the number of times the object type occurs in that region out of the total number of times the object type appears in the dataset. For example, we see that on average, similar number of chairs are found in dining room and lobby (circle radius), but overall chairs appear more frequently in the dining room in the dataset (color).
}
\label{fig:obj_to_region_plot}
\end{figure}

We also provide statistics of object support relations to architecture elements (\Cref{fig:obj_to_arch_support_plot}) and other objects (\Cref{fig:obj_to_obj_support_plot}).  As expected, we see that chairs typically go on floors, while curtains are supported by walls.
From \Cref{fig:obj_to_obj_support_plot}, we see that cushions are typically found on beds, chairs, and couches while towels are typically found on shelves.
Similarly, in \Cref{fig:obj_to_region_plot} we show the object-to-region statistics.  We see that some object categories tend to appear more frequently in a particular region (i.e. room) type. For example, couches are more frequently found in living rooms than in bedrooms.


\clearpage
\newpage

\begin{figure*}[t]
\includegraphics[width=\linewidth]{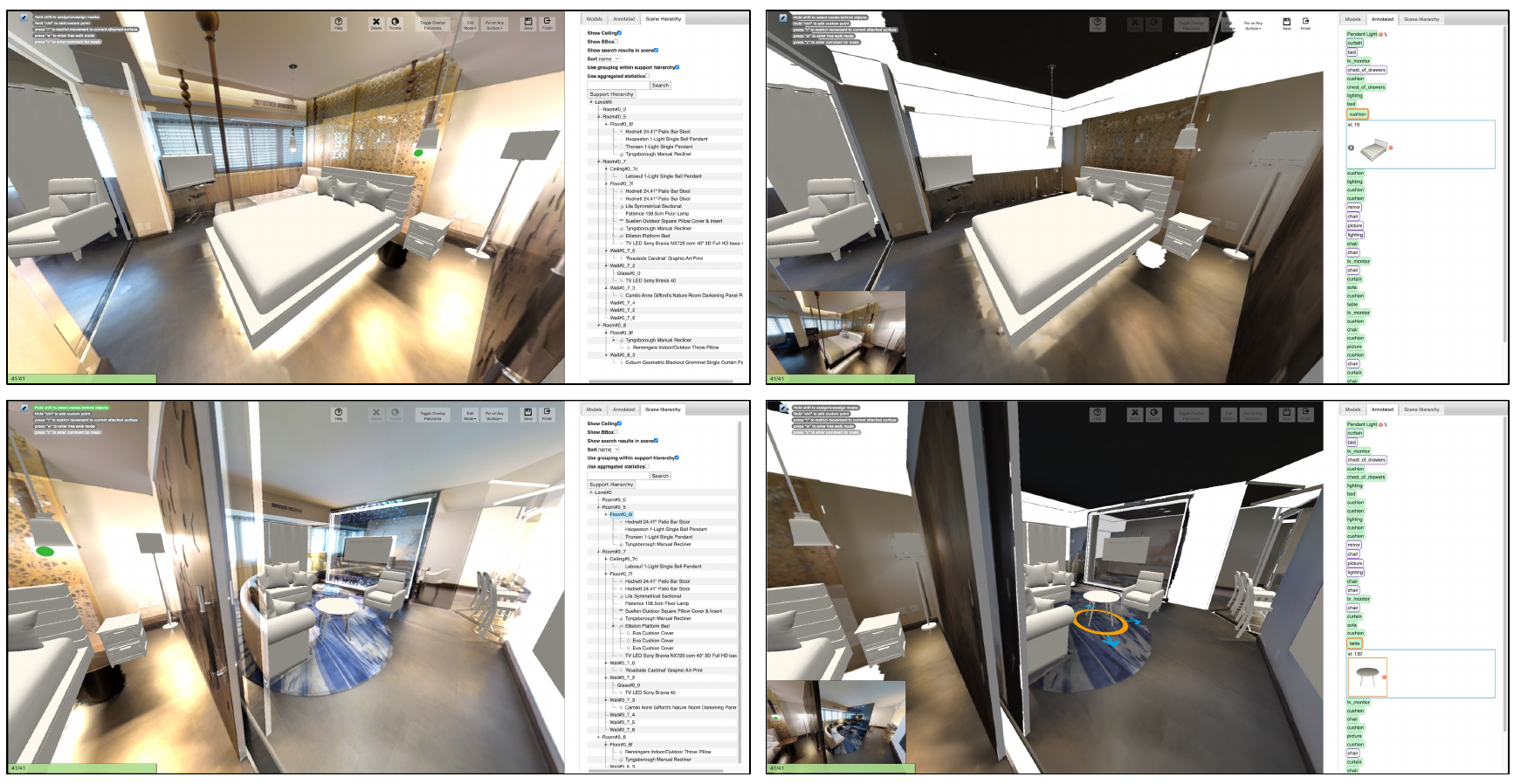}
\caption{
The \dataset interface allows users to select 3D CAD models and place them in the scene.  Users can see the alignment of the object against the panorama. 
Camera controls allows the user to rotate the camera and to see different perspective views of the scene.
The user can also toggle the overlaid panorama on (left) or off (right), to get a better view of the underlying 3D architecture and all objects placed into the scene thus far using the interface.
}
\label{fig:rlsd_interface}
\end{figure*}

\begin{figure}[t]
\centering
\includegraphics[width=\linewidth]{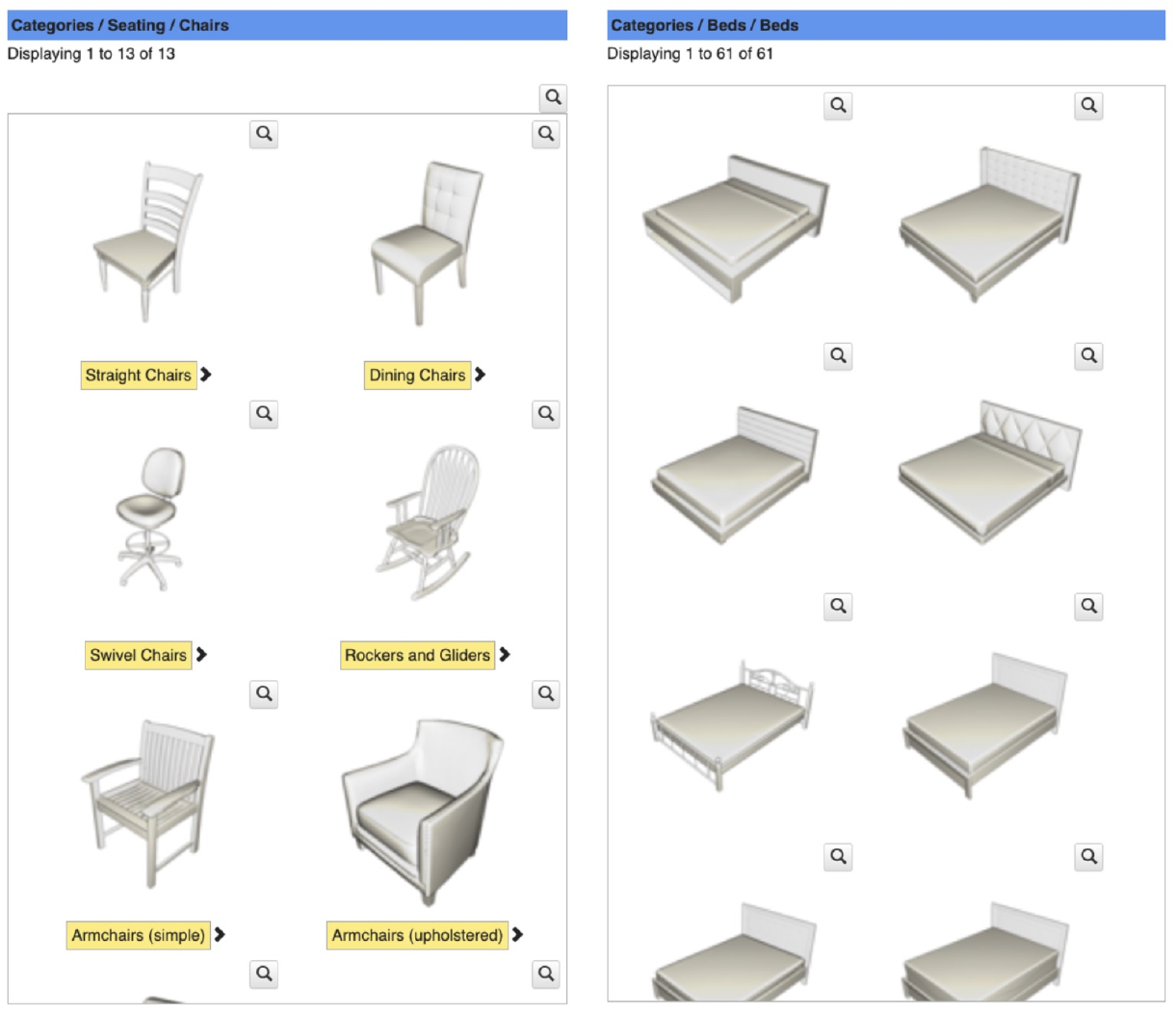}
\caption{
The user sees a list of candidate CAD models that is filtered depending on the semantic object category of the mask that was clicked in the panorama view.
The list is hierarchical, allowing the user to refine the category into finer-grained categories such as the examples of chair types on the left side, and then select an appropriate instance of a chair within the finer category.
}
\label{fig:interface_model_selection}
\end{figure}

\begin{figure}[t]
\centering
\includegraphics[width=0.8\linewidth]{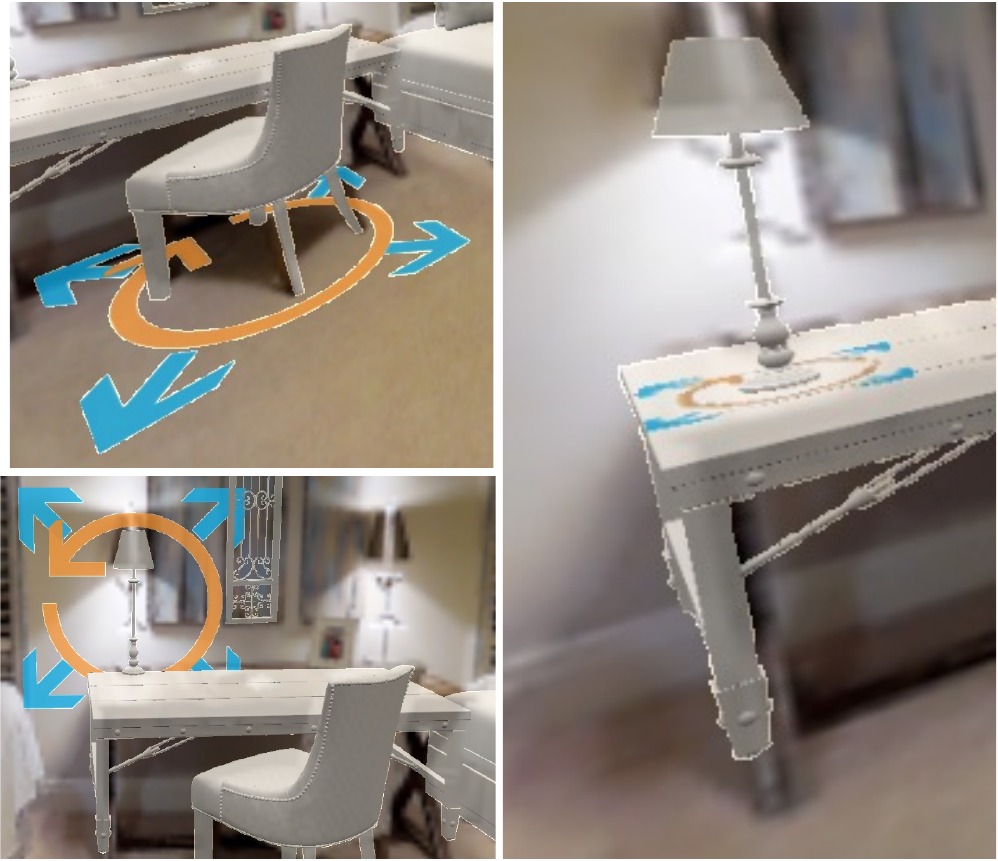}
\caption{
The interface allows annotators to attach objects to either architecture or other objects. The \textcolor{orange}{manipulator} is oriented on the attached surface and allows for rotating the object whereas the \textcolor{cyan}{arrows} allow for scaling.
}
\label{fig:object_attachment}
\end{figure}

\section{Annotation interface details}
\label{sec:supp-annotation-interface}
Our annotation interface consists of a web interface developed using \texttt{\href{https://threejs.org}{three.js}} that allows users to insert 3D assets into the scene while \emph{visually overlaid} on the panorama.
To achieve this, our interface assumes that there is a set of panoramas with corresponding camera poses, and an 3D architecture on which the objects can be placed.  
We implement two viewing modes, panorama mode and architecture mode, to let users switch between overlaid panorama and underlying 3D scene. 

\mypara{Data Assets.}
We construct a parametric 3D architecture for 20 Matterport3D scenes.
We take the region annotations that specify wall segments to create the initial 3D architecture.
We then  project annotations for the labels relating to windows and doors to get an initial estimate for the placement of doors and windows on the architecture.
Next, we create a textured architecture by rendering the reconstructed scene onto the estimated surfaces of each architecture element plane (wall, floor, ceiling).
Using a 3D interface that shows the architecture, we manually refining the wall boundaries and the placement of doors and windows on the walls to correct any prominent errors.
The projection of door and window annotations onto the walls is often noisy due to open doors, inaccurate windows, and noises in the annotation.
We obtain each RGB panorama by stitching 6 skybox images from the same camera viewpoint.
During the data preprocessing stage, we also parse panoramas into semantic object instance masks to provide reference objects during annotation.
We get these instance masks by rendering segmentations from Matterport3D's annotated object instance house meshes. 

\begin{figure}
\includegraphics[width=\linewidth]{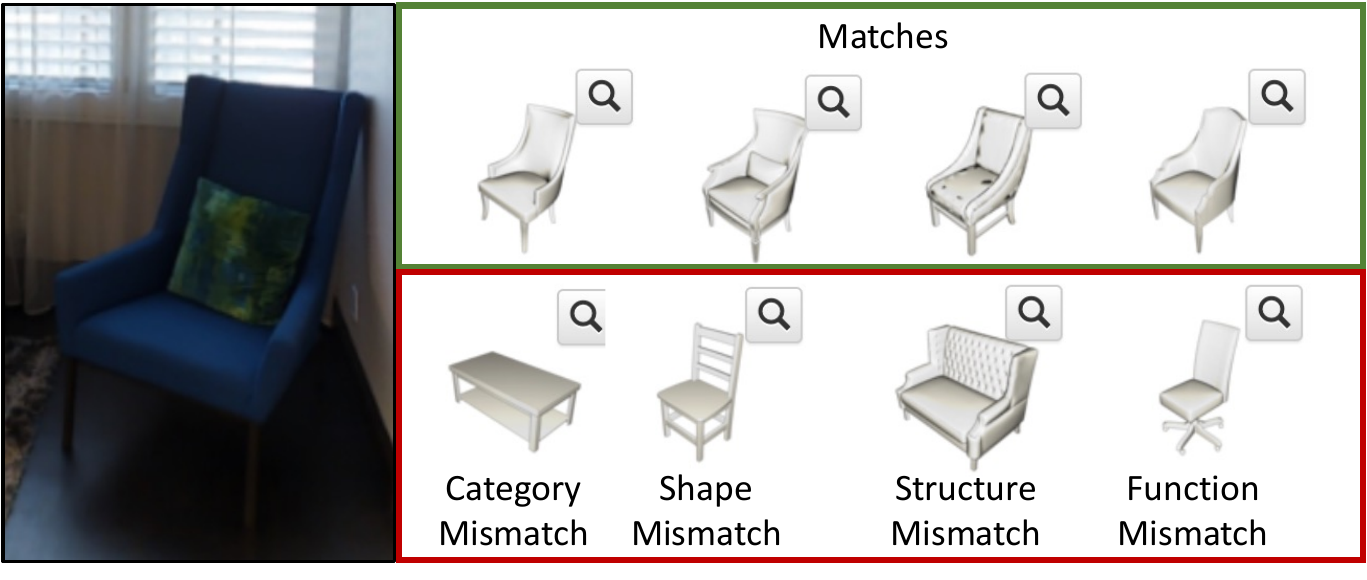}
\caption{
\textbf{Object selection in \dataset.} Annotators are instructed to select objects to insert in to a scene based on how well they match with the object observed in the panoramic image. Top: shows good object matches that an annotator would select following our instructions. Bottom: shows different types of mismatching objects that annotators are instructed to avoid.
}
\label{fig:obj_similarity}
\end{figure}

\mypara{Annotation process.}
We describe a typical annotation workflow starting with an empty scene (see \Cref{fig:rlsd_interface}).
A user freely pans the camera to explore the whole scene while the overlay is kept in sync. 
After clicking an object to be annotated in the panorama, a list of candidate 3D shapes of the same category is shown in a side panel (see \Cref{fig:interface_model_selection}). 
The user is instructed to identify the best matching 3D shape (see \Cref{fig:obj_similarity}).  
The inserted 3D shape is automatically placed at the location in the scene where the user initially clicked. The user can further manipulate the position, scale, and orientation of objects so that the object is aligned to the image (see \Cref{fig:object_attachment}). 
The placement is attached to a specific surface already in the scene, thus creating a scene support hierarchy by construction.

We recruited annotators and instructed them to follow these guidelines:
1) \textit{Completeness}: each mask should be annotated with a 3D model of an object. Some masks may be divided into parts for different objects and some masks may be merged into one (see detailed discussion of ``mask-to-object assignment''). If an important object does not have a mask, it can still be added (see discussion of ``custom masks'').
2) \textit{Object match}: the categories, shapes and sizes of the placed objects match those observed (see ``object selection'' criteria)
3) \textit{Spatial accuracy}: object placements and orientations should be as close to those observed in the panorama (see ``object selection'' criteria). There should be no collisions or floating objects. 

\begin{figure}[t]
\centering
\includegraphics[width=0.9\linewidth]{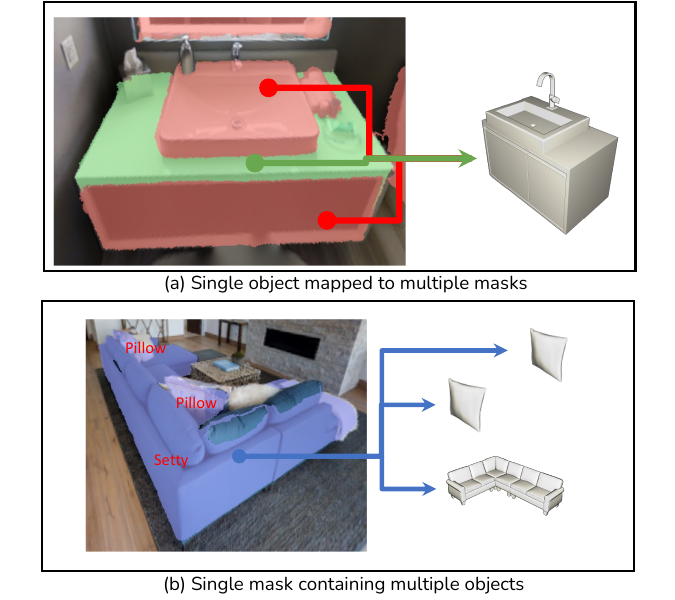}
\caption{
\textbf{Mask-to-object assignment.}
Our annotation strategy allows for objects that need to be assigned to multiple instance masks (e.g., sink at top), and multiple masks needing to be assigned to the same object (e.g., couch and pillows at bottom).
}
\label{fig:rlsd_many2many_feat}
\end{figure}
\mypara{Mask-to-Object Assignment.}
In some cases, it is overly restrictive to assume that there is a one-to-one correspondence between masks and objects.
For example, an object may need to be assigned to multiple masks because the two masks correspond to parts of the same object, separated by occlusion.
In other cases, we have masks that include multiple objects (see \Cref{fig:rlsd_many2many_feat}).
Our system supports these cases such that a user can place multiple models for the same mask by re-selecting a mask that already has a model assigned and inserting an additional model. For cases where a model is shared among multiple masks, the user first inserts the model having selected one of the masks. Then, the user can assign other relevant masks to the already inserted model.
Handling of these cases enables us to correctly annotate densely cluttered arrangements such as kitchen cabinetry, sink units, and pillows on couches.

\mypara{Object Selection.} 
We decompose the requirement on semantically-matching objects into 4 sub-aspects (see \Cref{fig:obj_similarity}): category, shape, structural, and functional similarity.
For example, a category mismatch constitutes a `chair' being annotated with a `table', a shape mismatch constitutes `high-back armchair' being annotated with a `dining-chair' model, a structural mismatch constitutes a `single-seater chair' being annotated with a `double-seater chair' and a functional mismatch constitutes a `an armchair with no wheels' being annotated with a `swivel chair with wheels and no arms' (\Cref{fig:obj_similarity}).
We exclude door and window objects for annotations since they are represented as holes on the walls of the architecture and their placement can be largely automated.

\begin{figure}
\includegraphics[width=\linewidth]{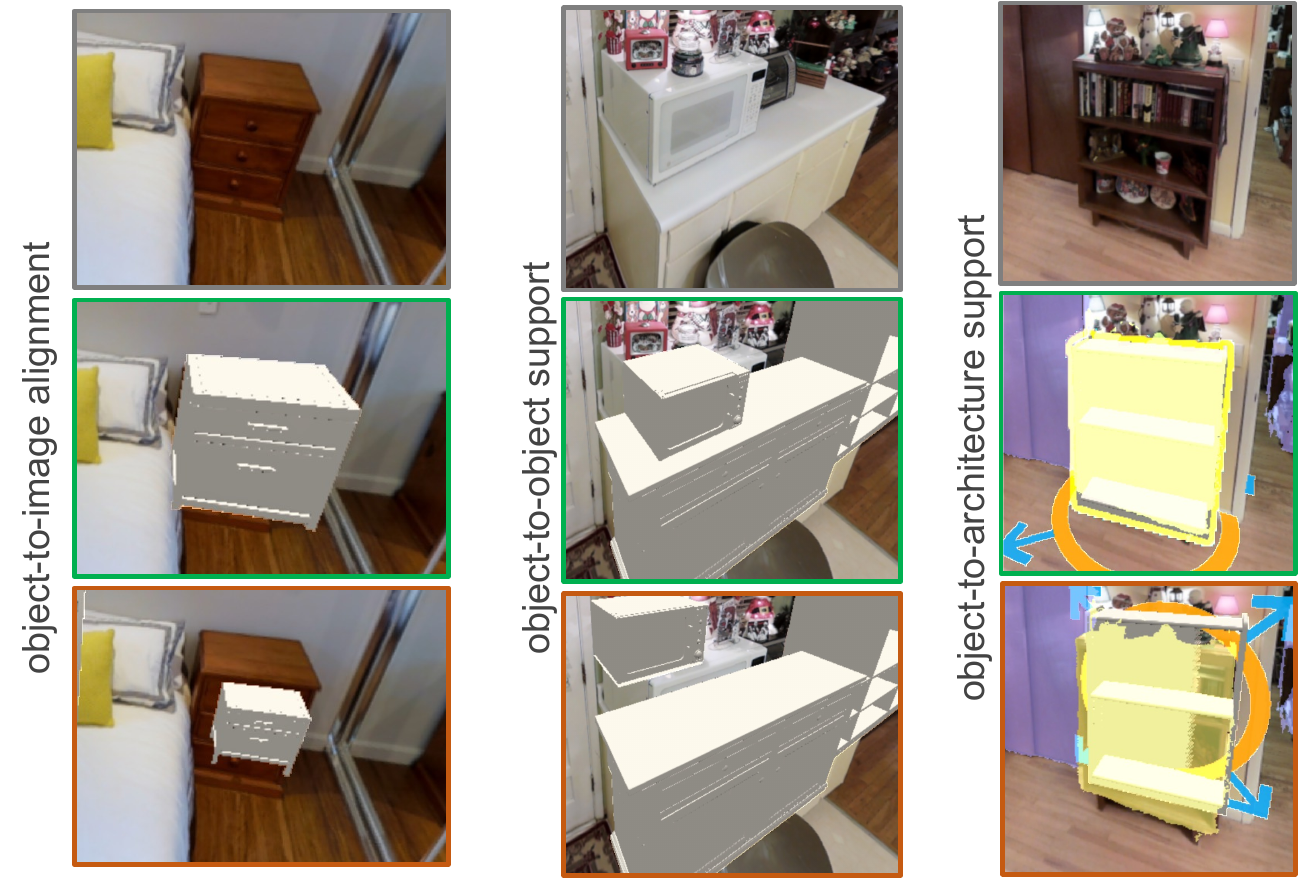}
\caption{
\textbf{Object alignment in \dataset.} (Left) Objects are closely aligned to the image. (Middle) Objects are properly supported by other objects. (Right) Objects are supported by appropriate architecture elements.
In each column, the top is a cropped view from the panorama, the middle highlighted in green is a correctly placed object, while the bottom is an example of an error that annotators avoid using our annotation interface.
}
\label{fig:obj_alignment}
\end{figure}

\begin{figure}
\includegraphics[width=\linewidth]{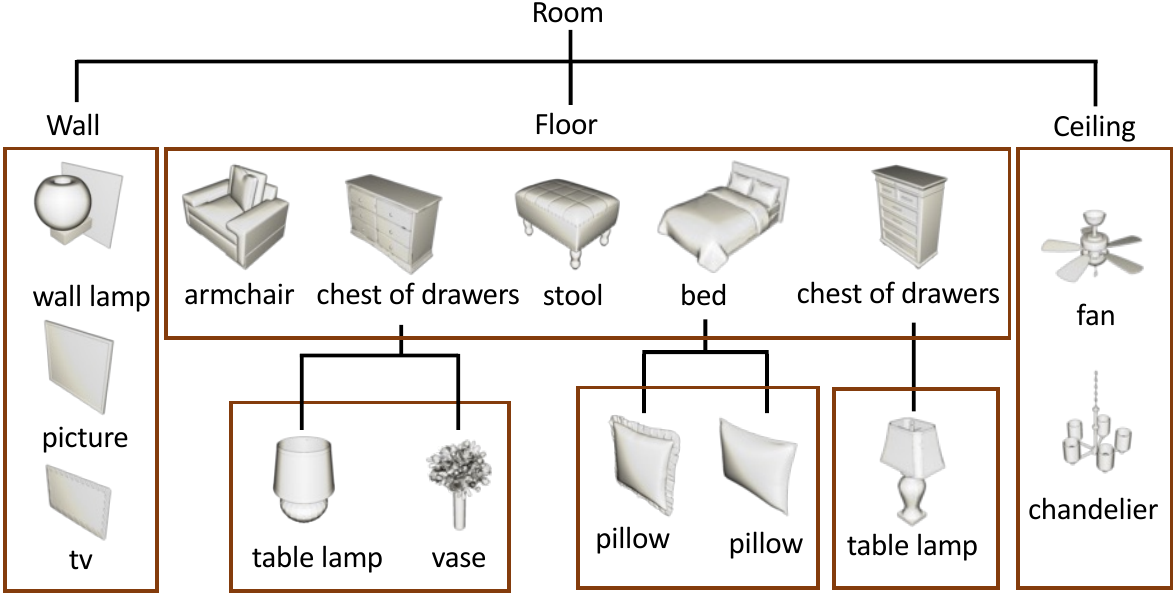}
\caption{
\textbf{Scene object support hierarchy.}
Objects in a scene are supported by either architecture elements or other objects.
}
\label{fig:scene_hierarchy}
\end{figure}

\mypara{Object alignment and support.} 
Additionally, the objects can have two types of support structure: i) object-to-object support; and ii) object-to-architecture support.
Object-to-object support ensures that two objects are supported by each other properly.
For example, a microwave placed on a counter is by construction constrained to be on the counter top, and not to float in midair.
Similarly, in the object-to-architecture support case, an object placed on an architectural element (floor, wall or ceiling) is ensured to be supported by the planar surface of that element.
This type of annotation also helps to disambiguate some otherwise physically implausible scenarios.
For example, a chest of drawers is typically supported by the floor, and not by the adjacent wall (\Cref{fig:obj_alignment}).
In \Cref{fig:obj_to_arch_support_plot}, we show a concrete example of how different objects are attached to architectural elements and supported by other objects.

\mypara{Custom Masks.}
We further allow annotators to insert objects for which there are no existing instance masks to ensure the scenes are densely populated and objects are properly supported. 
In some cases, the user may decide to leave a mask unannotated.
This could be because the mask is invalid or there are no viable models for the object.
In this case, the user can mark the object as `unannotated' and leave comments explaining the reason.





\end{document}